%% file: main.tex
\documentclass[conference]{IEEEtran}
\IEEEoverridecommandlockouts
\usepackage{cite}
\usepackage{amsmath,amssymb,amsfonts}
\usepackage{algorithmic}
\usepackage{graphicx}
\usepackage{xspace}
\usepackage[table]{xcolor}
\usepackage{amsmath}
\usepackage{balance}
\usepackage[ruled,linesnumbered]{algorithm2e}
\usepackage{rotating}
\usepackage{makecell}
\usepackage{multirow}
\usepackage{textcomp}
\usepackage[hyphens,spaces,obeyspaces]{url}
\usepackage{xcolor}
\usepackage{subcaption}
\usepackage{algorithm2e}
\usepackage{soul}
\usepackage{hyperref}
\usepackage{pifont}

\urldef{\footurl}\url{https://www.dropbox.com/scl/fo/carrcdl9v13b2813e58ui/h?rlkey=vakpjh83xt2ui6o8r51xljm32&dl=0}

\urldef{\footurlnew}\url{https://www.dropbox.com/scl/fi/lq01y9qbbzbvaqnh7owva/SiloFuse_appendix.pdf?rlkey=ed0bf2lb8pmc9g4siey665s3b&dl=0}

\newcommand{\para}[1]{\vspace{1mm}\noindent\textbf{#1}.} 
\newcommand{\parab}[1]{\vspace{1mm}\noindent\textbf{#1}}

\def\BibTeX{{\rm B\kern-.05em{\sc i\kern-.025em b}\kern-.08em
    T\kern-.1667em\lower.7ex\hbox{E}\kern-.125emX}}
\begin{document}

\title{
{\algo: Cross-silo Synthetic Data Generation with Latent Tabular Diffusion Models}\\}

\author{
\IEEEauthorblockN{Aditya Shankar}
\IEEEauthorblockA{
\textit{Distributed Systems}\\
\textit{TU Delft}\\
Delft, The Netherlands\\
{a.shankar@tudelft.nl}
\and
\IEEEauthorblockN{Hans Brouwer}
\IEEEauthorblockA{
\textit{BlueGen.ai}\\
The Hague, The Netherlands\\
hans@bluegen.ai}
\and
\IEEEauthorblockN{Rihan Hai}
\IEEEauthorblockA{
\textit{Web Information Systems} \\
\textit{TU Delft}\\
Delft, The Netherlands\\
{r.hai@tudelft.nl}}
\and
\IEEEauthorblockN{Lydia Chen}
\IEEEauthorblockA{
\textit{Dept. of Computer Science} \\
\textit{University of Neuchatel/ TU Delft}\\
Neuchatel, Switzerland\\
{lydiachen@ieee.org}}}}

\newtheorem{exmp}{Example}[section]
\newcommand{\revision}[1]{{\color{black}#1}}
\newcommand{\new}[1]{{\color{blue} #1}}
\newcommand{\algol}{\texttt{LatentDiff}\xspace}
\newcommand{\algo}{\texttt{SiloFuse}\xspace}
\newcommand{\galgol}{\texttt{GAN(linear)}\xspace}
\newcommand{\galgoc}{\texttt{GAN(conv)}\xspace}
\newcommand{\algotab}{\texttt{TabDDPM}\xspace}
\newcommand{\algoe}{\texttt{E2E}\xspace}
\newcommand{\algoed}{\texttt{E2EDistr}\xspace}
\newcommand{\lc}[1]{{\color{magenta}\texttt{} Lydia: #1 }} 
\newcommand{\as}[1]{{\color{teal}\texttt{} Aditya: #1 }} 
\newcommand{\rh}[1]{{\color{blue}\texttt{} Rihan: #1 }} 
\newcommand{\hb}[1]{{\color{orange}\texttt{} Hans: #1 }} 
\newcommand{\norm}[1]{\left\lVert#1\right\rVert}

\newcommand*\colourcheck[1]{%
  \expandafter\newcommand\csname #1check\endcsname{\textcolor{#1}{\ding{52}}}%
}
\newcommand*\colourcross[1]{%
  \expandafter\newcommand\csname #1cross\endcsname{\textcolor{#1}{\ding{54}}}%
}
\colourcheck{orange}
\colourcross{red}
\colourcheck{teal}

\maketitle

\begin{abstract}

Synthetic tabular data is crucial for sharing and augmenting data across silos, especially for enterprises with proprietary data. However, existing synthesizers are designed for centrally stored data. Hence, they struggle with real-world scenarios where features are distributed across multiple silos, necessitating on-premise data storage. We introduce \algo, a novel generative framework for high-quality synthesis from cross-silo tabular data. To ensure privacy, \algo utilizes a distributed latent tabular diffusion architecture. Through autoencoders, latent representations are learned for each client's features, masking their actual values. We employ stacked distributed training to improve communication efficiency, reducing the number of rounds to a single step. Under \algo, we prove the impossibility of data reconstruction for vertically partitioned synthesis and quantify privacy risks through three attacks using our benchmark framework. Experimental results on nine datasets showcase \algo's competence against centralized diffusion-based synthesizers. Notably, \algo achieves 43.8 and 29.8 higher percentage points over GANs in resemblance and utility. Experiments on communication show stacked training's fixed cost compared to the growing costs of end-to-end training as the number of training iterations increases. Additionally, \algo proves robust to feature permutations and varying numbers of clients.

\end{abstract}

\input{icde_vfl_difuss/introduction}

\input{icde_vfl_difuss/background}
\input{icde_vfl_difuss/methodology}

\input{icde_vfl_difuss/experiments}

\input{icde_vfl_difuss/relatedwork}

\input{icde_vfl_difuss/futureworkandconclusion}
\newpage




\end{document}


\def\BibTeX{{\rm B\kern-.05em{\sc i\kern-.025em b}\kern-.08em
    T\kern-.1667em\lower.7ex\hbox{E}\kern-.125emX}}

\title{
{Appendix: \algo}\\}

\newcommand{\algo}{\texttt{SiloFuse}\xspace}
\newcommand{\algol}{\texttt{LatDiff}\xspace}
\newcommand{\algotab}{\texttt{TabDDPM}\xspace}
\newcommand{\algoe}{\texttt{E2E}\xspace}
\newcommand{\algoed}{\texttt{E2EDistr}\xspace}
\newcommand{\ra}[1]{{\color{magenta}\texttt{} Reviewer 1: #1 }} 
\newcommand{\rb}[1]{{\color{teal}\texttt{} Reviewer 2: #1 }} 
\newcommand{\rc}[1]{{\color{blue}\texttt{} Reviewer 3: #1 }} 
\newcommand{\meta}[1]{{\color{orange}\texttt{} Meta reviewer: #1 }} 
\newcommand{\norm}[1]{\left\lVert#1\right\rVert}

\maketitle
\section{Capture of tabular features} 

\begin{figure}[htb]
    \centering
    \begin{subfigure}{0.225\textwidth}
        \includegraphics[width=\textwidth]{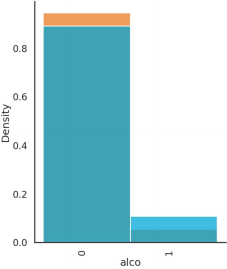}
        \caption{\algo Cardio - Discrete}
    \end{subfigure}
    \begin{subfigure}{0.225\textwidth}
        \includegraphics[width=\textwidth]{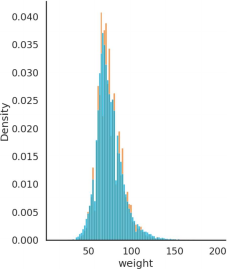}
        \caption{\algo Cardio-Continuous}
    \end{subfigure}
    \begin{subfigure}{0.225\textwidth}
        \includegraphics[width=\textwidth]{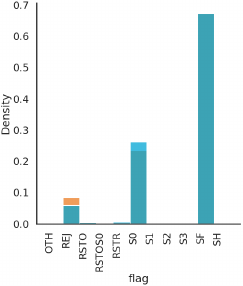}
        \caption{\algo Intrusion - Discrete}
    \end{subfigure}
    \begin{subfigure}{0.225\textwidth}
        \includegraphics[width=\textwidth]{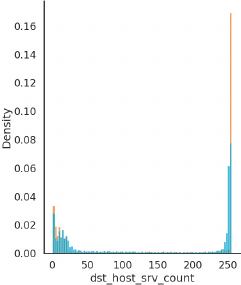}
        \caption{\algo Intrusion - Continuous}
    \end{subfigure}

    \begin{subfigure}{0.225\textwidth}
        \includegraphics[width=\textwidth]{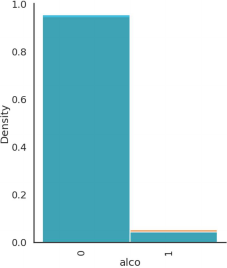}
        \caption{\algotab Cardio - Discrete}
    \end{subfigure}
    \begin{subfigure}{0.225\textwidth}
        \includegraphics[width=\textwidth]{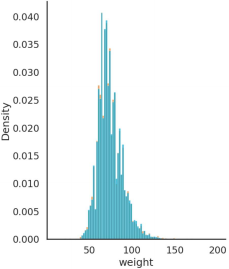}
        \caption{\algotab Cardio - Continuous}
    \end{subfigure}
    \begin{subfigure}{0.225\textwidth}
        \includegraphics[width=\textwidth]{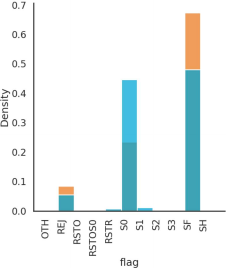}
        \caption{\algotab Intrusion - Discrete}
    \end{subfigure}
    \begin{subfigure}{0.225\textwidth}
        \includegraphics[width=\textwidth]{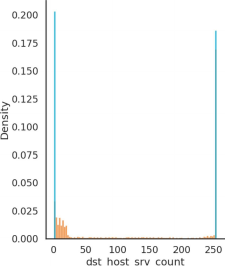}
        \caption{\algotab Intrusion - Continuous}
    \end{subfigure}
    
\caption{Feature distributions of categorical and continuous features captured by \algo and \algotab. \textbf{Real data in orange and synthetic in blue.}} 
\label{fig:featuredist}
\end{figure}
 We show the feature distributions for the real and synthetic data on two datasets for \algo and \algotab in \autoref{fig:featuredist}. This tests whether latent models can capture the heterogeneous distributions of tabular data. Four clients were used for \algo.

The figure shows the feature distributions of \algo and \algotab for one discrete and one continuous feature. We observe that \algo can capture both discrete and continuous features in tabular data. On the simpler dataset, Cardio, we see that \algotab can better capture the categorical feature distributions than \algo. This may be because the low sparsity in Cardio allows \algotab to achieve higher performance through the multinomial loss. On the continuous feature, \algotab performs only slightly better.

 However, we see a change in the more challenging dataset (Intrusion). Due to the high sparsity of discrete features, \algo performs better than \algotab for both continuous and discrete features. Here, converting into continuous latents boosts \algo by reducing the feature sparsity introduced by one-hot encoding, allowing the model to capture the relationships better.

%% file: icde_vfl_difuss/introduction.tex
\section{Introduction}


Today's enterprises hold proprietary business-sensitive data and seek collaborative solutions for knowledge discovery while safeguarding privacy. For example, cardiologists and psychiatrists gather patients' heart rates and mental stress levels, respectively, for potential joint treatments~\cite{de2022intriguing}. However, privacy regulations like GDPR~\cite{GDPR2016} restrict sharing such \emph{cross-silo} datasets, i.e., \textit{feature}-partitioned or \textit{vertically}-partitioned datasets, across enterprises. Yet, their importance in data management beckons novel solutions to learn over vertically-partitioned data silos~\cite{vaidya2002privacy,vaidya,navathe}.

\begin{figure}[tb]
    \centering
    \includegraphics[width=0.5\textwidth]{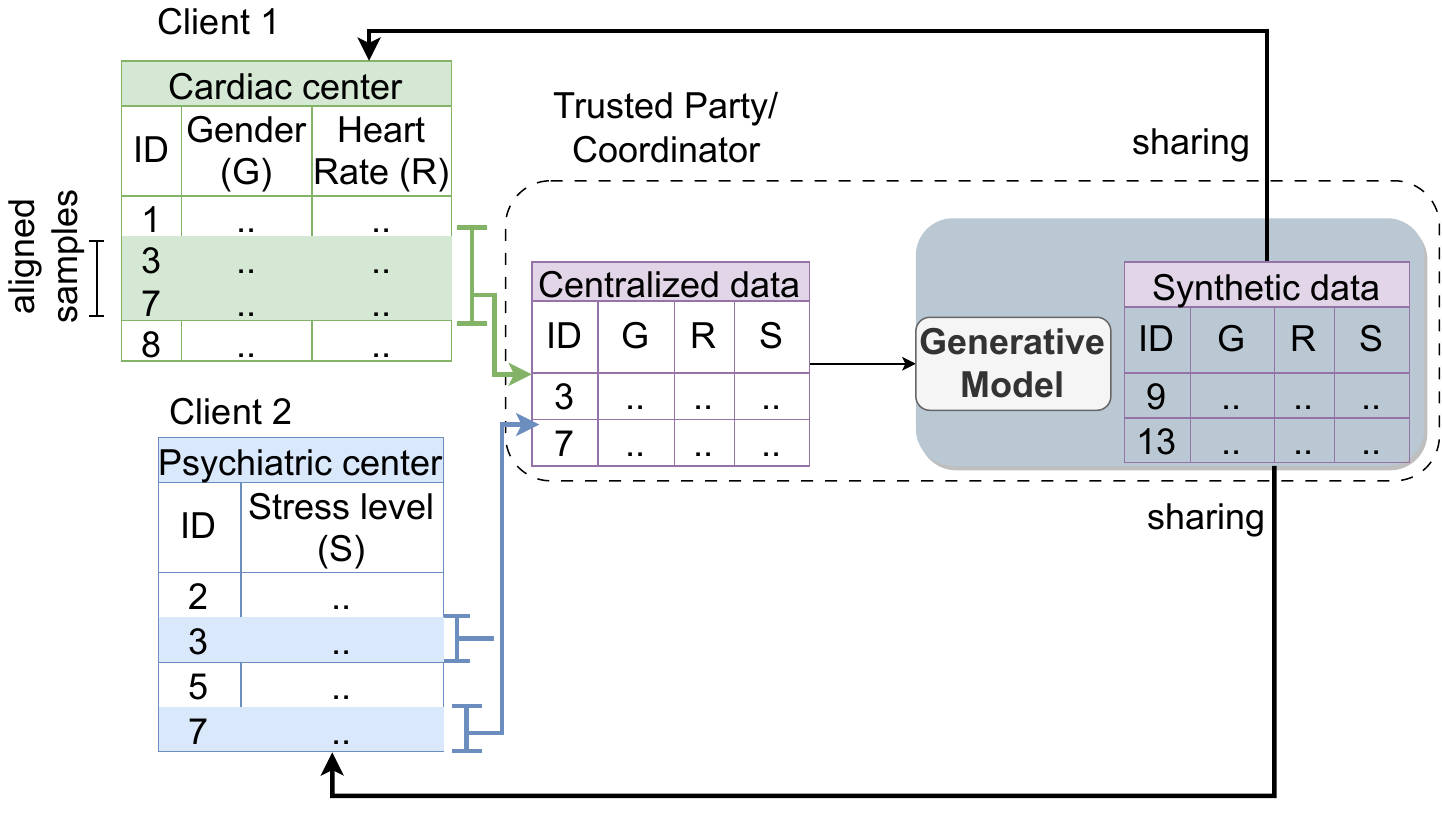}
    \caption{Synthetic data for data sharing and augmentation. Features from a cardiac center (client 1) and a psychiatric center (client 2) are centralized from the common patient IDs and a generative model synthesizes new samples to share with the clients, augmenting their datasets and enabling collaboration without sharing real data.
    }
    \label{fig:tradsoln}
\end{figure}

In this regard, the database community is looking towards using synthetic data as an alternative to real data for protecting privacy~\cite{icdesynth,sigmodsynth,vldbsurvey,tablegan}. The current tabular generative models encompass technologies ranging from autoencoders~\cite{choi2017generating, oord2017neural}, flow-based models~\cite{lee2022differentially}, autoregressive (AR) models~\cite{mahiou2022dpart}, score-based models~\cite{kim2022stasy}, Generative Adversarial Networks (GANs)~\cite{tablegan,xu2019modeling, wang2023differentially,zhao2023gtv,zhao2021ctab}, and the state-of-the-art \textit{diffusion} models~\cite{kotelnikov2023tabddpm,ho2020denoising,hoogeboom2021argmax}. Recent findings show that diffusion models excel over previous technologies, including the dominant GAN-based approaches~\cite{kotelnikov2023tabddpm,dhariwal2021diffusion}. This may stem from GANs' training instability, resulting in mode-collapse and lower sample diversity compared to diffusion models~\cite{bayat2023study}. Despite their promise, extending diffusion models to cross-silo data requires \textit{centralizing} data at a trusted party before synthesis, as shown in Fig.~\ref{fig:tradsoln}. This violates the constraint of having data on-premise, defeating the purpose of privacy preservation. 

Cross-silo methods using federated learning~\cite{li2020federated,konevcny2016federated} already exist for GANs~\cite{zhao2023gtv,wang2023differentially}. These are based on centralized models, such as Table-GAN~\cite{tablegan}, CTGAN~\cite{xu2019modeling}, and CTAB-GAN~\cite{zhao2021ctab}. However, diffusion-based cross-silo solutions do not exist despite their promising results over GANs. Therefore, we tackle the research problem of designing and training high-quality tabular diffusion synthesizers for cross-silo data.

Developing a cross-silo tabular synthesizer poses several challenges. First, tabular data has a mix of \textit{categorical} and \textit{continuous} features that require encoding for training. The mainstream \textit{one-hot encoding}\cite{kotelnikov2023tabddpm,hoogeboom2021argmax} for categorical variables increases the difficulty in modeling distributions due to \textit{expanded feature sizes} and \textit{data sparsity}, and provides \textit{poor obfuscation} to sensitive features. Second, for synthetic data to resemble the original distributions, capturing cross-silo feature correlations is needed. However, learning the global feature correlation is challenging without having access to a centralized dataset. Third, training on data spread across silos is expensive due to the \textit{high communication costs} of distributed training. Traditional methods, such as \textit{model parallelism}\cite{dean2012large,vepakomma2018split}, split the model across multiple clients/machines, incurring high communication overhead due to the repeated exchange of forward activations and gradients between clients.


 Given these challenges, we design, \algo, a novel framework with a tabular synthesizer architecture and an efficient distributed training algorithm for feature-partitioned data. Motivated by the recent breakthrough of diffusion models in generating high-quality synthetic data using \textit{latent} encodings~\cite{rombach2022high}, the core of \algo is a \textit{latent tabular synthesizer}. First, autoencoders encode sensitive features into continuous latent features. A generative Gaussian diffusion model~\cite{ho2020denoising} then learns to create new synthetic latents based on the latent embeddings of the original inputs. By merging latent embeddings during training, the generative model learns global feature correlations in the latent space. These correlations are maintained when decoding latents back into the real space. As a result, \algo generates new synthetic features while keeping the original data on-premise.

 \algo incorporates multiple novelties. First is the architecture and training paradigm for the tabular synthesizer. Autoencoders and the diffusion model are trained separately in a \textit{stacked} fashion, with the latent embeddings being communicated to the diffusion backbone only once. This keeps communication rounds at a \textit{constant} regardless of the number of training iterations for the generator and the autoencoders. Second, our design allows the synthetic data to be generated while retaining the vertical partitioning. This has stronger privacy guarantees compared to existing cross-silo generative schemes that centralize synthetic data~\cite{zhao2023gtv,wang2023differentially}. Third, we further develop a \textit{benchmark framework} to evaluate \algo and other baselines regarding the synthetic data quality and its utility on downstream tasks. Theoretical guarantees show the impossibility of data reconstruction under vertically partitioned synthesis, and risks are quantified when synthetic data is shared. Extensive evaluation on nine data sets shows that \algo is competitive against centralized methods while efficiently scaling with the number of training iterations.


\para{Contributions} 
\algo is a novel distributed framework for training latent tabular diffusion models on cross-silo data. It's novel features make the following contributions. 
\begin{itemize}
    \item A \textbf{latent tabular diffusion model} that combines autoencoders and latent diffusion models. We unify the discrete and continuous tabular features into a shared continuous latent space. The backbone latent diffusion model captures the feature correlations across silos by centralizing the latents. 
    
    \item A \textbf{stacked training paradigm} that trains local autoencoders at the clients in parallel, followed by latent diffusion model training at the coordinator/server. Decoupling the training of the two components lowers communication to a single round, overcoming the high costs of end-to-end training. 
    \item A \textbf{benchmarking framework} that computes a resemblance score by combining five statistical measures and a utility score by comparing the downstream task performance. We also prove the impossibility of data reconstruction when the synthetic data is kept vertically partitioned and quantify the privacy risks of centralizing synthetic data using three attacks.

\end{itemize}

%% file: icde_vfl_difuss/background.tex
\section{Background and problem definition}
  
In this section, we commence with the background on tabular diffusion models, focussing on the current state-of-the-art centralized synthesizer, \algotab \cite{kotelnikov2023tabddpm}. We then formally define the problem setting for cross-silo tabular synthesis, identify existing research gaps, and elucidate the limitations of current centralized methods in addressing this challenge.
 
\subsection{Background: Diffusion models for tabular data synthesis}
\label{subsec:ddpm}

\begin{figure}[bt]
    \centering
    \includegraphics[width=0.4\textwidth]{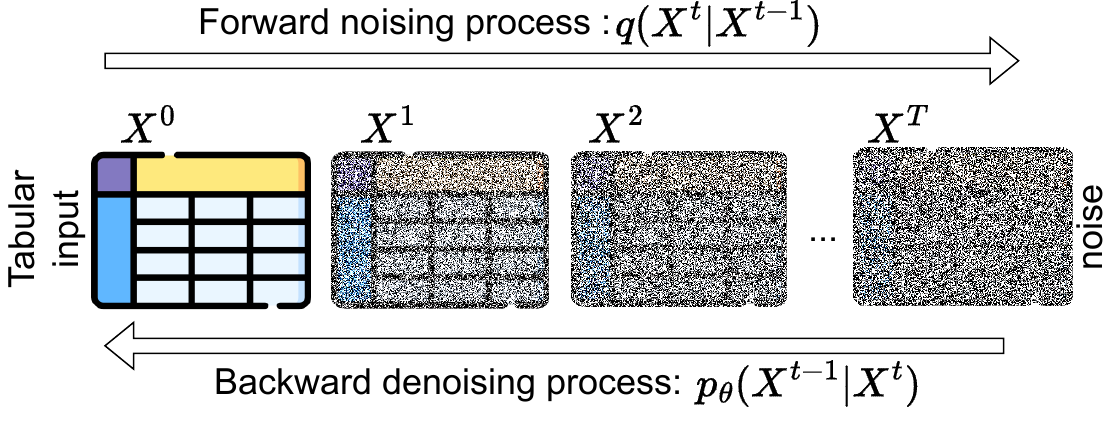}
 
    \caption{Forward and backward diffusion process. $X^0$ is an input dataset of samples and $X^T$ is the fully-noised dataset after $T$ timesteps.}
    \label{fig:diffusionprocess}
\end{figure}

\textit{Denoising diffusion probabilistic models} (DDPMs) generate data by modeling a series of forward (noising) and backward (denoising) steps, as a Markov process~\cite{ho2020denoising}. Fig.~\ref{fig:diffusionprocess} depicts tabular synthesis with this scheme. A dataset $X^0  \in \mathbb{R}^{n \times d}$, has $n$ samples $\{x_1^0,x_2^0,..,x_n^0\}$ and $d$ features. It produces a noisy dataset $X^T$ over $T$ noising steps. The reverse process iteratively denoises $X^T$ to obtain a clean dataset $X^0$. 

As tabular data contains both \emph{continuous} and \emph{categorical} features, it requires different encoding and noising/denoising methods, which we describe below. \autoref{tab:notations} summarises the mathematical notations used.
 
 \para{Continuous features} The forward noising transitions, $q(X^t|X^{t-1})$, as shown in Fig.~\ref{fig:diffusionprocess}, gradually noise the dataset over several steps. Each transition is modeled as a normal distribution, using a variance schedule with fixed constants  $\beta^t$ $\forall t \in [1:T]$~\cite{ho2020denoising}. Through a reparameterization trick mentioned in Ho et al.~\cite{ho2020denoising}, the forward transition $q(X^t|X^0)$ for a sequence of $t$ steps can be done in a single step:
\begin{equation}
     F(X^0,t,\epsilon) = q(X^t|X^0) = \sqrt{\bar{\alpha}^t}X^0 + \sqrt{1-\bar{\alpha}^t}\epsilon
 \end{equation}
 Here, the function $F$ obtains the noisy samples $X^t$ after $t$ forward steps,  $\bar{\alpha}^t = \prod_{j=1}^t(1-\beta^j)$, and $\epsilon \sim \mathcal{N}(0,I)$ is a base noise level.  

 The reverse transitions $p_\theta(X^{t-1}|X^t)$, are modelled using a neural network with parameters $\theta$. It learns these denoising transitions by estimating the noise added, quantified using the mean-squared-error (MSE) loss~\cite{ho2020denoising}:
\begin{equation}
     \mathcal{L}^t=\mathop{\mathbb{E}_{X^0,\epsilon,t}}
     \norm{\epsilon - \epsilon_{\theta}(X^t,t)}_2^2
     \label{eq:smalldiff}
 \end{equation}
 Here, $\epsilon_{\theta}(X^t,t)$ predicts the base noise from the noised features at step $t$. By minimizing this loss, the neural network reduces the difference between the original inputs and the denoised outputs. Predicting the added noise brings the denoised outputs closer to the actual inputs. Hence, \eqref{eq:smalldiff} can be viewed as minimizing the mean-squared-error (MSE) loss between the inputs and denoised outputs.
 
 \para{Discrete features} 
 Working with categorical and discrete features requires \emph{Multinomial DDPMs}~\cite{hoogeboom2021argmax}. Suppose the dataset's $v$-th feature, i.e., $X^t[v]\in \{0,1\}^{n \times k}$, is a categorical feature with a one-hot embedding over $k$ choices. Adding noise over its available categories changes this feature during the forward process. At each step, the model either picks the previous choice $X^{t-1}[v]$ or randomly among the available categories, based on the constants  $\beta^t$~\cite{hoogeboom2021argmax}. This way, it gradually evolves while considering its history.
The multinomial loss, denoted as $\mathcal{M}^{t}[v]$, measures the difference between ($X^t[v]$) and $X^{t-1}[v]$, in terms of the Kullback-Leibler divergence. Averaging the losses over all categorical features produces the final loss.


The combined loss of \algotab from the continuous  and categorical losses is as follows:
\begin{equation}  \mathcal{L}^t_{\algotab} = \mathcal{L}^t + \sum_{v \in V}\mathcal{M}^t[v]/|V| \label{eq:tabddpmloss}
\end{equation}
Here, $V$ denotes the index set of the categorical features.

\begin{figure}
    \centering
    \includegraphics[width=0.4\textwidth]{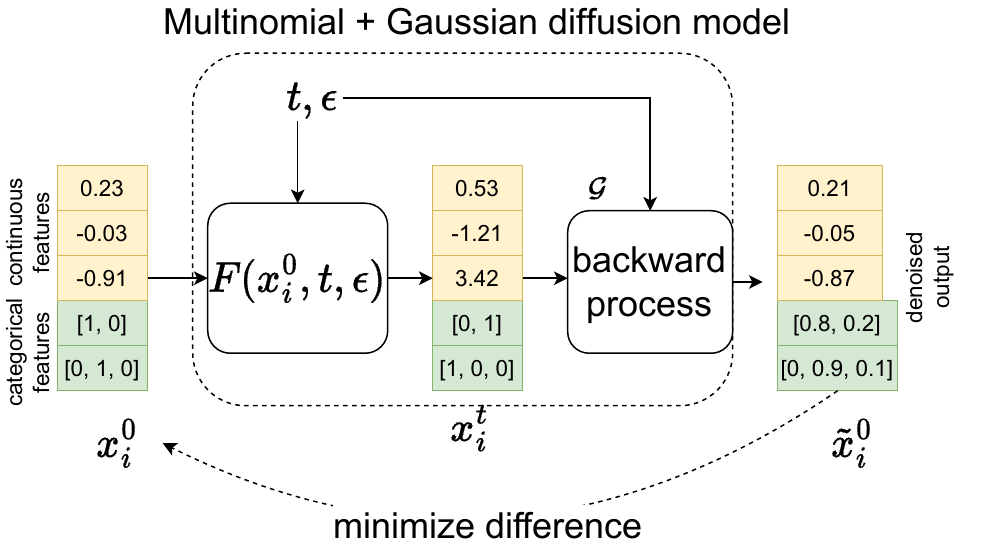}
    \caption{Design of \algotab. Internally, the model predicts the noise added to $x_i^t$ to minimize the difference with the original inputs after denoising.}
    \label{fig:tabddpm}
    \vspace{-0.4cm}
\end{figure}

\begin{exmp}
\label{exmp:tabddpmexmp}
Consider Fig.~\ref{fig:tabddpm}, where discrete features for a sample $x_i^0$ such as gender (M/F) and marital status (single, divorced, married) are one-hot encoded as $[1, 0]$ and $[0, 1, 0]$ respectively. This encoding expands the feature size of a single discrete column to 2 and 3 columns, respectively. Other continuous features with values 0.23, -0.03, and -0.91 may also be present. Using a combination of multinomial and continuous (Gaussian) DDPMs, the model noises the inputs using forward process $F$ and then denoises (using backbone $\mathcal{G}$) over multiple timesteps $t$, giving outputs $\tilde{x}_i^0$. Intuitively, the model minimizes the difference between the reconstructed output $\tilde{x}_i^0$ and the original inputs $x_i^0$ by predicting the added noise. Mathematically, this combines the multinomial and diffusion losses using \eqref{eq:tabddpmloss}. 
\end{exmp}
\vspace{-0.1cm}	

\subsection{Problem Definition: Cross-silo tabular synthesis}
\para{Input} In the considered scenario, there exist \emph{M} distinct parties or clients $\{C_1, ..., C_M\}$. Each party $C_i$ possesses a subset of features $X_i \in \mathbb{R}^{n\times d_i}$, where $n$ is the total number of samples and $d_i$ is the number of features owned by client $C_i$. These originate from a feature-partitioned dataset $X \in \mathbb{R}^{n\times d}$, with $n$ samples and $d=\sum_{i=1}^M d_i$ features that is spread across multiple silos: $X = X_1||X_2||...||X_M$, where $||$ represents column-wise concatenation. All the features $X_{i \in {1:M}}$ are assumed to have aligned samples (rows) with other clients. This means only the rows corresponding to the common samples are selected, as shown in Fig.~\ref{fig:tradsoln}. Private-set intersection schemes achieve this through a common feature such as a sample ID~\cite{hardy2017private,scannapieco2007privacy}. Without loss of generality, we assume party $C_1$ is the \textbf\textit{{coordinator}} and holds the generative diffusion model $\mathcal{G}$. 

\para{Objective} The primary objective is to generate a synthetic dataset $\tilde{X} \in \mathbb{R}^{n\times d} = \tilde{X_1}||\tilde{X_2}||...||\tilde{X_M}$ whose distribution is close to that of the original features while maintaining the privacy of their actual values. Each client $C_{1\leq i\leq M}$, possesses the partition $\tilde{X_i} \in \mathbb{R}^{n \times d_i}$ with synthetic features corresponding to their respective features while maintaining the sample alignment across clients. After generation, the feature sets $\tilde{X}_{i \in {1:M}}$ can be kept vertically partitioned or shared with other clients for downstream tasks.

\para{Constraints and Assumptions} Due to the sensitive nature of each client's original features, they are considered confidential and, therefore, cannot be directly shared with other participants in plaintext. In other words, the original feature data that is vertically partitioned stays on the client's premise. Moreover, we assume all parties are \emph{honest-but-curious or semi-honest}~\cite{hardy2017private}. This implies that all participants (including the server) follow the training and synthesizing protocols, but can attempt to infer others' private features solely using their own data and any information communicated to them. Also, malicious behavior, like violating the protocols or feeding false data, is prohibited.


\subsection{Research gaps}
\label{subsec:researchgapsbckg}
Several reasons make the cross-silo scenario challenging to centralized tabular DDPM methods. 

First, methods such as \algotab~\cite{kotelnikov2023tabddpm} and multinomial diffusions~\cite{hoogeboom2021argmax} require categorical features to be \textit{one-hot encoded}. This induces sparsity and increases feature sizes, as explained in Example~\autoref{exmp:tabddpmexmp} and Fig.~\ref{fig:tabddpm}. This can lead to complications, as the increased feature complexity increases the chances of overfitting. 


Second, naively extending such methods using model-parallism~\cite{dean2012large} incurs high communication costs. For methods such as \algotab to be amenable to cross-silo dataset scenarios, clients must encode their local features into latents, which are then sent to the coordinator holding the generative backbone $\mathcal{G}$. As seen in Fig.~\ref{fig:tabddpm}, this would involve sending one-hot encodings to the backbone model held by the coordinator, which becomes more expensive due to the feature size expansion. We later show the increase in feature sizes due to one hot encoding in \autoref{tab:DataStatistics}, under \hyperref[sec:experiments]{Section V}.
Moreover, training requires end-to-end backpropagation with multiple communication round-trips between the coordinator and the clients. This scales poorly with increasing training iterations as the clients and the coordinator must exchange gradients and forward activations for every iteration. 

Third, from the privacy front, centralized tabular synthesizers do not consider the risks associated with sharing synthetic features with other parties in cross-silo scenarios. As we explain in Example \autoref{exmp:privacy}, links between synthetic features could allow parties to model the dependencies and reconstruct private features of the other parties, posing a risk. Additionally, implementing end-to-end training in a distributed setting requires communicating gradients across parties. This increases susceptibility to attacks that exploit gradient-leakage for inferring private data~\cite{jin2021cafe,zhu2019deep,geiping2020inverting}.

 \begin{exmp}
 \label{exmp:privacy}
 Consider a scenario where Company A possesses personal information about individuals, such as names and addresses. At the same time, Company B has individuals' financial data, such as income and spending habits. By sharing synthetic features, an adversary might detect links between certain financial behaviors and specific names or addresses. The attacker could infer or deduce patterns linking financial behaviors to specific individuals or households as synthetic features could unintentionally mirror the actual data.
 \end{exmp}

%% file: icde_vfl_difuss/methodology.tex
\section{\algo} 
This section presents our framework \algo, with a distributed architecture for latent tabular DDPMs. We first explain the intuition and novelty of our framework, followed by an overview of the architecture, training, synthesis, and privacy analysis. 

\begin{figure}[t]
    \centering
    \includegraphics[width=0.4\textwidth]{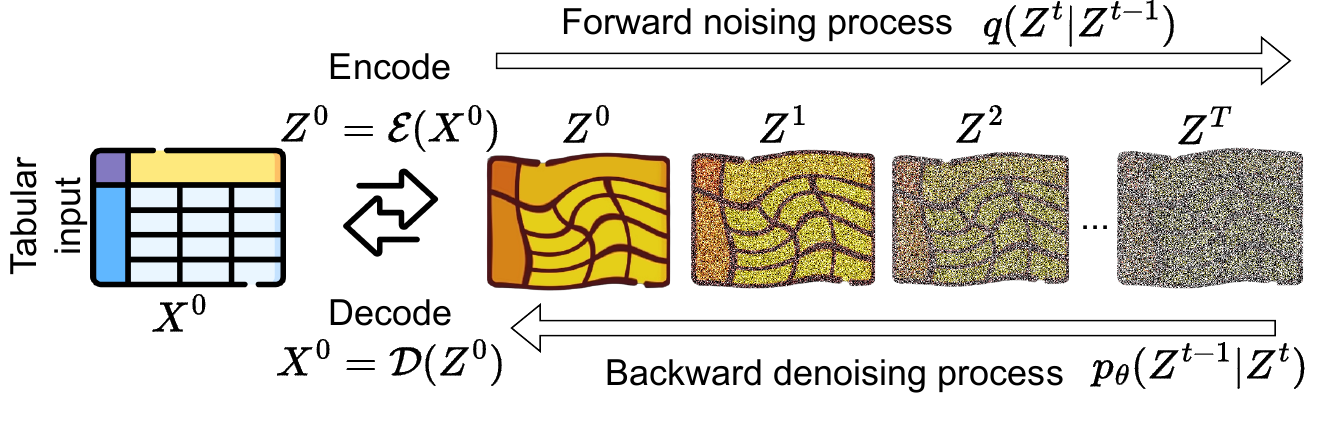}
    \caption{Centralized latent tabular DDPM. $Z$ denotes the latent space.}
    \label{fig:latentdiffprocess}
    \vspace{-3mm}
\end{figure}

\algo trains latent DDPMs \cite{rombach2022high}, using autoencoders to encode real features into latent embeddings. The coordinator concatenates these to generate synthetic latents using the DDPM backbone (illustrated in Fig.~\ref{fig:latentdiffprocess}). Local decoders at the clients then transform the outputs to the real space without the coordinator accessing the real features.
\input{icde_vfl_difuss/tables/notation}

This allows for the novelty of decoupling the training of autoencoders and the DDPM, capturing cross-silo feature correlations within the latent space without real data leaving the silos. 
\begin{figure} [t]
    \centering
    \includegraphics[width=0.45\textwidth]{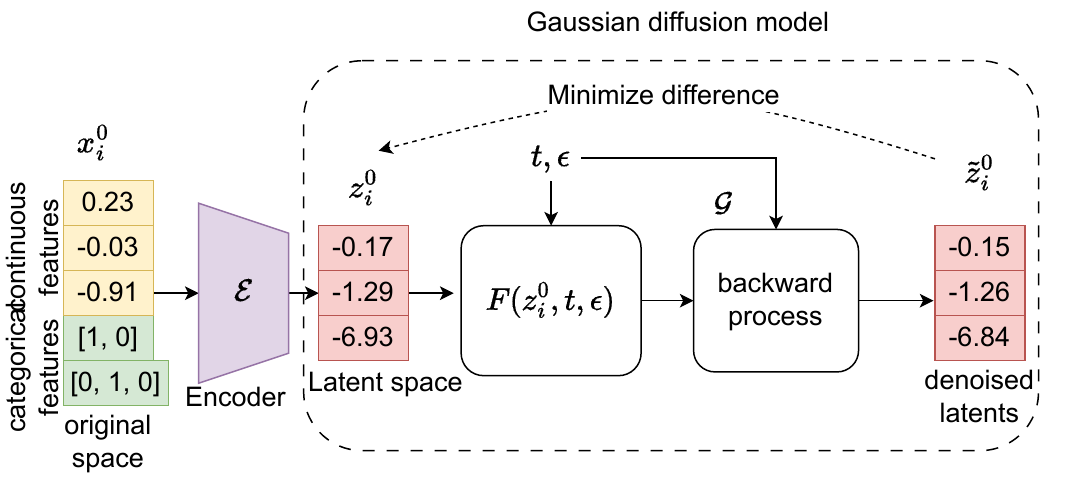}
    \caption{Centralized latent DDPM. While $\mathcal{E}$ is the encoder, there is also a decoder $\mathcal{D}$ (not shown) for translating latents back to the original space. Internally, the DDPM predicts the noise added during the forward process, which minimizes the gap between $z_i^0$ and $\tilde{z}_i^0$}
    \label{fig:latentdiffarch}
    \vspace{-0.3cm}
\end{figure}

\begin{figure*}[t]
    \centering
    \begin{subfigure}{0.35\textwidth}
        \includegraphics[width=\textwidth]{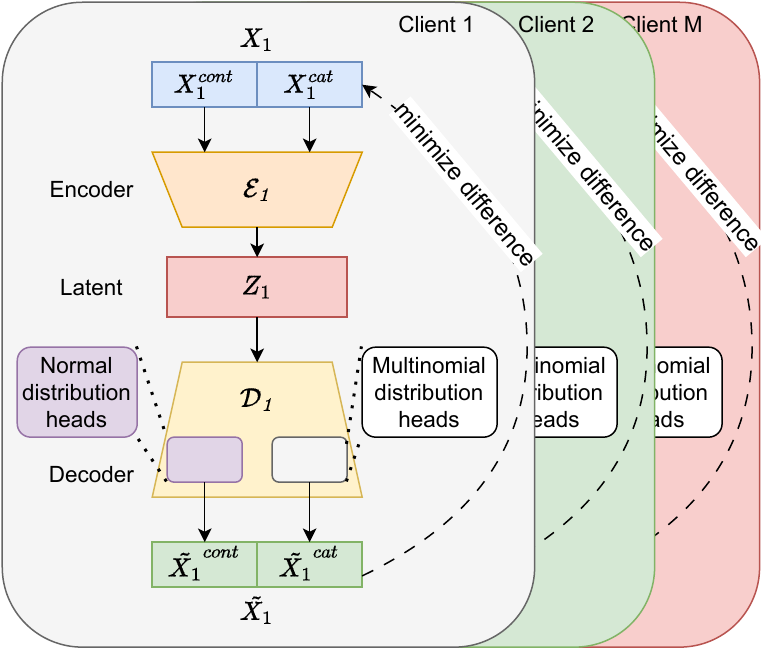}
        \caption{Step 1: Local autoencoder training}
        \label{fig:localautoenc}
    \end{subfigure}
    \hspace{16mm}
    \begin{subfigure}{0.35\textwidth}
        \includegraphics[width=\textwidth]{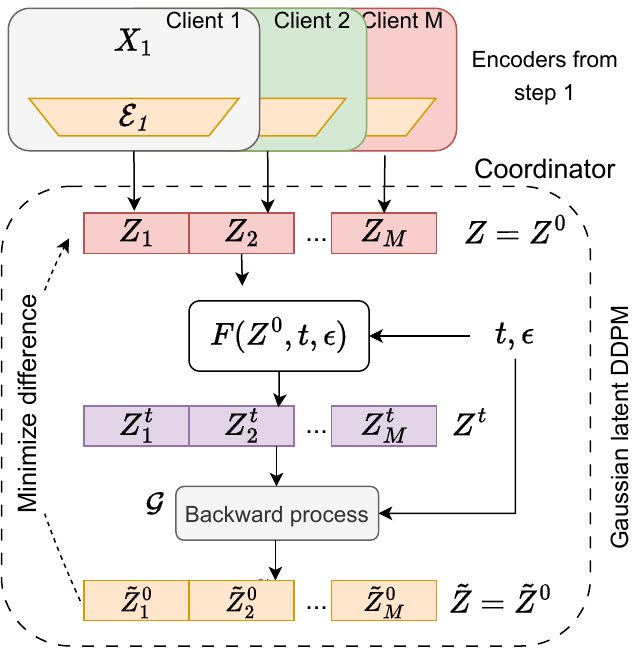}
        \caption{Step 2: Coordinator trains latent DDPM}
        \label{fig: latentdiff}
    \end{subfigure}
    \caption{Two step training in \algo. $X_i^{cont}$ and $X_i^{cat}$ correspond to the continuous and categorical features.}

    \label{fig:twostep}
\end{figure*}

\subsection{Why the shift to latent space?}

Latent DDPMs have several advantages over operating on the real space directly. Firstly, they use encoders to convert both categorical and continuous features into continuous latents, reducing sparsity and size compared to one-hot encodings (Fig.~\ref{fig:latentdiffarch}). This is quantitatively shown later in \autoref{tab:DataStatistics}, with one-hot encoding increasing feature sizes by nearly 10 to 200 times on some datasets. Secondly, the training phases of the autoencoder(s) and the diffusion model are decoupled/stacked, thus reducing communication costs to just one round. In contrast, the end-to-end distributed scheme jointly trains both components, increasing the cost with the number of iterations.

\algo has a two-step training process. For a centralized dataset $X$, an encoder $\mathcal{E}$ transforms $X$ into a latent representation $Z = \mathcal{E}(X)$, where $Z \in \mathbb{R}^{n\times s}$ ($s$ represents the latent space's feature size). Subsequently, a decoder $\mathcal{D}$ minimizes the reconstruction loss to find optimal parameters $(\mathcal{E^*,D^*})$ for the autoencoder.

\begin{equation}
(\mathcal{E^*,D^*}) = \arg \min_{(\mathcal{E,D})} \mathcal{L}^{AE}(X, \mathcal{D}(\mathcal{E}(X)))
\label{eq:aeloss}
\end{equation}

Here, the loss function $\mathcal{L}^{AE}$ is chosen as negative log-likelihood or KL divergence. 
 The coordinator then trains the diffusion model $\mathcal{G}$ on the latents. The objective function is similar to \eqref{eq:smalldiff} to minimize the gap between the denoised outputs and the original inputs. Hence, the function simplifies to the MSE loss between the input latents $Z = Z^0$ and the noisy latents $Z^t$:

\begin{equation}
\mathcal{L}_{\mathcal{G}}^t=\mathop{\mathbb{E}_{Z=\mathcal{E}(X),t}}
\norm{Z - \mathcal{G}(Z^t,t)}_2^2
\label{eq:smalldiff2}
\end{equation}

Here, $\mathcal{G}(Z^t, t)$ returns the denoised version of the samples using the backward process. The noisy latents $Z^t$, are computed using the forward process $F(Z^0, t, \epsilon)$, as described in \hyperref[subsec:ddpm]{Section II-A}.  


\subsection{\algo  overview: Distributed latent diffusion}
\para{Architecture and functionalities} The model architecture of \algo is depicted in Fig.~\ref{fig:twostep}. 
Each client $C_{i \in \{1:M\}}$, trains a local autoencoder, i.e., encoder-decoder pair ($\mathcal{E}_i, \mathcal{D}_i$), which are privately held.
The coordinator holds the generative diffusion model $\mathcal{G}$. Without loss of generality, we assign this role to $C_1$. 

The true features, $X_i$, consist of categorical features $X_i^{cat}$ and continuous features $X_i^{cont}$, as shown in Fig.~\ref{fig:localautoenc}. The encoders convert these features to continuous latents: $Z_{i} \in \mathbb{R}^{n \times s_i} = \mathcal{E}_i(X_i)$ $\forall i \in \{1:M\}$. Like the original space, the latent dimensions, $s_i$, sum up to give the total latent dimension, i.e., $s=\sum_{i=1}^M s_i$. The decoders then convert these latents features back into the original space, i.e., $\tilde{X_i} = \mathcal{D}(Z_i)$ $\forall i \in \{1:M\}$. Similar to tabular \textit{variational autoencoders} (VAEs)~\cite{oord2017neural,xu2019modeling}, each decoder's head outputs a probability distribution for each feature. A typical choice for continuous features, $X_i^{cont}$, is the \textit{Gaussian distribution} where the head outputs the mean and variance parameters to represent the spread of feature values.

A multinomial distribution head is used for categorical or discrete features, $X_i^{cat}$, giving probabilities for the different categories or classes. 

 After training the autoencoders, the encoded latents ($Z = Z_1||Z_2||...||Z_M$) centralize at the coordinator (see Fig.~\ref{fig: latentdiff}). These undergo forward noising via function $F$, resulting in $Z^t = F(Z^0, t, \epsilon)$. The backward denoising produces $\tilde{Z}=\tilde{Z}^0=\tilde{Z_1}||\tilde{Z_2}||...||\tilde{Z_M}$. Training a standard Gaussian DDPM on the continuous latents ($Z$) using MSE \eqref{eq:smalldiff2}, eliminates the need for a separate multinomial loss. Centralizing latents instead of actual features enables the model to learn cross-silo feature correlations within the latent space, which are preserved in the original space upon decoding.
 
\begin{figure}[tb]
    \centering
    \includegraphics[width=0.35\textwidth]{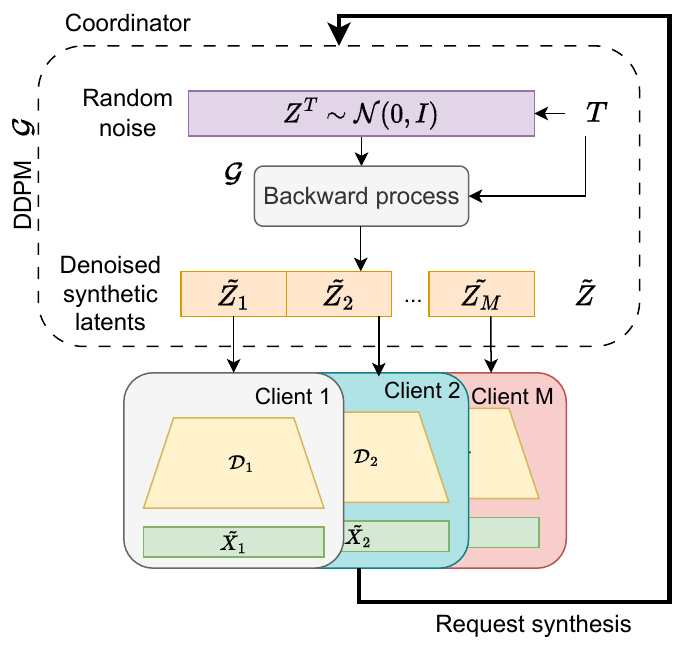}
    \caption{Sample generation by stacking the DDPM and decoders. A requesting client, $C_2$, triggers synthesis.}
    \label{fig:inference}
\end{figure}

 In synthesis, a client requests the coordinator to synthesize new samples (refer to Fig.~\ref{fig:inference}). The coordinator begins by sampling random noise $Z^T$. This is then denoised over several timesteps $T$, producing $\tilde{Z} = \tilde{Z_1}||\tilde{Z_2}||...||\tilde{Z_M}$. Each client, $C_i$, receives a part of the synthetic latents, i.e., $\tilde{Z_i}$. Using their private decoder $\mathcal{D}_i$, clients convert these into the original space, i.e.,  $\tilde{X_i} = \mathcal{D}_i(\tilde{Z_i})$. This generates feature-partitioned synthetic samples while preserving inter-feature associations.




\section{\algo: Distributed Training and Synthesis}

In this section, we describe the algorithms for training and synthesis and discuss the privacy implications in \algo. We provide theoretical arguments for privacy based on latent irreversibility for vertically partitioned synthesis. We also indicate the risks of sharing features post-generation. 

\subsection{Distributed Training}
\label{subsec:autoenctraining}
\algo's training process adopts a stacked two-step approach, departing from standard end-to-end training. Fig.~\ref{fig:twostep} illustrates the two-step procedure detailed in \hyperref[alg:twosteptraining]{Algorithm 1}.

\textbf{Individual Autoencoder Training} (lines [\ref{ln:1.1}-\ref{ln:1.8}] of \hyperref[alg:twosteptraining]{Algorithm~1}):
For input sample features $X_{i \in {1:M}}$, client $C_i$ converts all $X_i$ features into numerical embeddings, employing one-hot encoding for categorical features. These continuous and categorical embeddings pass through an MLP of an encoder ($\mathcal{E}_i$), producing continuous-valued latents $Z_i$. A  decoder processes these latents, using output heads to map each feature to a normal or categorical distribution akin to other tabular VAEs~\cite{oord2017neural,xu2019modeling}.   The loss function \eqref{eq:aeloss}, measures the logarithm of the probability density between the output distributions and input features, for both categorical and discrete features.

\RestyleAlgo{ruled}

\SetKwComment{Comment}{/* }{ */}

\textbf{Generative Diffusion Model Training}:
After training the autoencoders, each client $C_{i \in \{1:M\}}$ produces latent features of the original training samples, i.e., $Z_i = \mathcal{E}(X_i)$, and communicates these to the coordinator. The coordinator concatenates these to obtain $Z = Z_1||...||Z_M$ ((lines \ref{ln:1.10},\ref{ln:1.12})).

The coordinator then trains locally for multiple epochs. It samples random base noise level $\epsilon$ and timesteps $t$ at each epoch and performs the forward noising process $F(Z^0, t, \epsilon)$. It then internally predicts the noise level added during the forward process to update the generator $\mathcal{G}$ for modeling the backward process (lines \ref{ln:1.13}-\ref{ln:1.17}). 

\subsection{Distributed Synthesis}
The DDPM backbone $\mathcal{G}$ and local decoders at each client are stacked for post-training synthesis, detailed in \hyperref[alg:inferencing]{Algorithm~2}.

Upon receiving a client's request, the coordinator samples Gaussian noise $Z^T$ (line \ref{ln:2.3}). Over multiple iterations in $\mathcal{G}$, noise is removed, yielding $\tilde{Z} = \mathcal{G}(Z^T, T)$. Clients then use their local decoder $\mathcal{D}_i$ to translate their partition of the latent samples $\tilde{Z_i}$ into real samples $\tilde{X_i} = \mathcal{D}_i(\tilde{Z_i})$. Links between $\tilde{Z_i}$ are preserved in $\tilde{X_i}$ due to the centralization of latents during training.


\subsection{Privacy in training}
The localized training of autoencoders and the diffusion model ensures that sensitive information remains confined within each client's domain. Despite sharing latent embeddings with the coordinator, the absence of decoders at the coordinator prevents data reconstruction from the latents. The coordinator aggregates latents to capture cross-client feature associations without observing the original features, thereby preserving privacy.
\begin{algorithm}[bt!]
\caption{Two-step training with \algo}
\label{alg:twosteptraining}
\KwData{Local feature sets $X_i$ on client $C_i$ $\forall k \in [1: M]$}
\KwResult{Trained autoencoders ($\mathcal{E}_i, \mathcal{D}_i$), max training timesteps $T$, and diffusion backbone $\mathcal{G}$}
\Comment{Local autoencoder training}
\nllabel{ln:1.1}\For{each client $C_i$ in parallel}{
    \For{$e_1$ epochs}{
        $Z_i \gets \mathcal{E}_i(X_i)$; \Comment{encoded latents}
        $\Tilde{X_i} \gets \mathcal{D}_i(Z_i)$; \Comment{Decoded latents}
        $Loss \gets \mathcal{L}^{AE}(\Tilde{X_i},X_i)$\Comment*[r]{\eqref{eq:aeloss}}
        Update $(\mathcal{E}_i, \mathcal{D}_i)$;
        
    }
}\nllabel{ln:1.8}
\For{each client $C_i$ in parallel}{
$Z_i \gets \mathcal{E}_i(X_i)$ \Comment*[r]{local latents}\nllabel{ln:1.10}
}
$Z = Z^0 \gets Z_1||Z_2||..||Z_i$ \Comment*[r]{coordinator}\nllabel{ln:1.12}
\For{$e_2$ epochs on coordinator}{
\nllabel{ln:1.13}$t, \epsilon \gets Uniform(\{1,T\})$, $\mathcal{N}(0,1)$\\
    $Z^t \gets F(Z^0, t, \epsilon)$ \Comment*[r]{forward process}
    $\tilde{Z} \gets \mathcal{G}(Z^t, t) $\Comment*[r]{Denoise}
\nllabel{ln:1.17}    Update $\mathcal{G}$ \Comment*[r]{\eqref{eq:smalldiff2}}
}

\end{algorithm}

\begin{algorithm}[bt!]
\caption{Synthesis with \algo}
\label{alg:inferencing}
\KwData{Trained decoders $\mathcal{D}_i$ on client $C_i$ $\forall i \in [1,M]$, backbone $\mathcal{G}$, requesting client $C_j$, and synthesis timesteps $T$}
\KwResult{Synthetic feature set $\Tilde{X_i}$ at $C_i$}
Client $C_j$ send request to coordinator\\

\textbf{At Coordinator:}\\
${Z^T} \gets \mathcal{N}(0,I)$ \Comment*[r]{Sample noise} \nllabel{ln:2.3}
$\tilde{Z} = \tilde{Z^0} \gets \mathcal{G}(Z^T, T)$ \Comment*[r]{Denoise}
Partition:
$\tilde{Z} = \Tilde{Z_1}||\Tilde{Z_2}||..||\Tilde{Z_M}$\\

\For{all clients $C_i$ in parallel}{
$\Tilde{X_i} \gets \mathcal{D}_i(\Tilde{Z_i})$ \Comment*[r]{Decode locally}
}
\Return{$\Tilde{X_i}$ on client $C_i$}
\end{algorithm}

Crucially, at no point during the training phase are raw, identifiable features shared. Instead, the communication revolves around derived latent representations or aggregated latent spaces, mitigating the risk of sensitive information exposure. Access to the decoders is necessary to revert latents to the real space. But as they are privately held, access is restricted, preventing the reversion process.


\subsection{Privacy in synthesizing} 
\label{subsec:privacysampling}
In our formulation, synthesis occurs in \textit{two} scenarios. In the first case, the synthesized data \textit{remains vertically partitioned}, with each client retaining only their respective features post-generation. In the second case, parties may share their synthetic features after generating them. While the latter allows parties to model downstream tasks independently, it introduces privacy risks due to links between features, as explained in example \autoref{exmp:privacy}. In contrast, the first case offers stronger guarantees by restricting access to other parties' features. However, it requires collaborative methods like \emph{vertical federated learning}~\cite{wei2022vertical,hardy2017private,liu2022vertical} for modeling downstream tasks, incurring a higher cost.

When features are shared, privacy guarantees become weaker, and we resort to empirical risk estimation, as shown in \hyperref[sec:empiricalprivacy]{Section V-$F$}. But for the first case, reconstructing inputs from just latents is challenging for the coordinator, as it is akin to finding the encoding function's inverse without knowing the inputs or the function itself. While intuitive, we formalize this here. Lemma 1 shows the impossibility of data reconstruction without knowing the domain of the inputs. Lemma 2 extends the impossibility result to cases where the domain is known but requires a domain with at least two elements. Both are building blocks for the \textit{latent irreversibility} theorem of \algo. 
\\\\
\textbf{Lemma 1:} For any function \( \mathcal{E}: X \rightarrow Z \), where \( Z \) is the codomain (latent space), there exists no oracle function \( \mathcal{O}: Z \rightarrow X \) capable of uniquely identifying the domain \( X \) (input space) without knowledge of the function \( \mathcal{E}: X \rightarrow Z \).
\\
\textbf{Proof:} For the sake of contradiction assume there is an oracle \( \mathcal{O}: Z \rightarrow X \) capable of identifying the domain \( X \) without knowledge of \( \mathcal{E} \). However, since there are infinitely many possible domains \( X \) that can map to a given codomain \( Z \), the oracle cannot uniquely map elements of \( Z \) to the correct domain.  
Therefore, such an oracle \( \mathcal{O} \) cannot exist. \(\blacksquare\)
\\\\
\textbf{Lemma 2:} For any surjective function \( \mathcal{E}: X \rightarrow Z \), where \( X \) has more than one element, there does not exist an oracle \( \mathcal{O}: Z \rightarrow X \) capable of uniquely identifying the pre-image\footnote{Pre-images with cardinality one: https://mathworld.wolfram.com/Pre-Image.html} of any element in \( Z \) without knowledge of \( \mathcal{E}: X \rightarrow Z \).
\\
\textbf{Proof:} Assume, for contradiction, the existence of an oracle \( \mathcal{O}: Z \rightarrow X \) capable of identifying the pre-image of any element in \( Z \) without knowledge of \( \mathcal{E} \). We consider two cases:
\begin{enumerate} 
   \item  If \( \mathcal{E} \) is not injective, then at least one element \( z \in Z \) has multiple pre-images in \( X \). This ambiguity makes it impossible for \( \mathcal{O} \) to uniquely identify a single pre-image for \( z \), contradicting the existence of \( \mathcal{O} \).
    %
    \item If \( \mathcal{E} \) is injective, we can construct another function \( \mathcal{E}': X \rightarrow Z \) by swapping the mappings of two distinct elements in \( X \) (i.e., \( \mathcal{E}'(x_1) = \mathcal{E}(x_2) \) and \( \mathcal{E}'(x_2) = \mathcal{E}(x_1) \)). This keeps the same codomain \( Z \) yet introduces ambiguity in determining the right pre-image without explicit knowledge of the function used, i.e., $\mathcal{E}'$ or $\mathcal{E}$. This scenario again contradicts the oracle's supposed capability to identify a unique pre-image.

\end{enumerate}
Therefore, such an oracle \( \mathcal{O} \) cannot exist. \(\blacksquare\)
\\\\
\textbf{Theorem 1 (Latent irreversibility)}:
In the \algo framework with private data (from input space $X$) and privately-held encoder and decoder functions ($\mathcal{E, D}$), the coordinator/server cannot reconstruct real samples from latent encodings alone when the domain $X$ is unknown. Reconstruction remains impossible if the domain is known and its size is greater than one.

\textbf{Proof:} Suppose, for contradiction, that the server can reconstruct real samples from just the latents. This implies the existence of an oracle $\mathcal{O}: Z \rightarrow X$, from the latent space $Z$ to the real space $X$.

We consider two cases based on Lemmas 1 and 2:
\begin{enumerate}
    \item Case 1:
Without knowing the encoding/decoding functions or the domain of $X$, the server cannot uniquely identify the real samples from the latents alone. In this case, Lemma 1 applies, so such an oracle $O$ cannot exist.
\item Case 2: If the domain $X$ is known, the server cannot reconstruct real samples from just the latents. In this case, Lemma 2 applies, so the oracle $O$ cannot exist. 

\end{enumerate}
Since both cases lead to contradictions, we conclude that the coordinator cannot reconstruct real samples from the latents alone in \algo's framework. $\blacksquare$

%% file: icde_vfl_difuss/tables/notation.tex
\begin{table}[tb]
    \centering
    \caption{Summary of key mathematical notations}
    \begin{tabular}{|c|c|}
    \hline
       Variable  &  Description\\\hline
        $q$ & Forward noising process\\
        $X,Z$ & Original dataset, latent dataset at step 0\\
        $X^t$, $Z^t$ & Dataset, latent dataset at $t$-th noising step\\
        $X^t[v]$ & $v$-th feature of dataset\\
        $x_i^t$, $\tilde{x}_i^t$ & Original and denoised $i$-th sample at timestep $t$\\
        $z_i^t$, $\tilde{z}_i^t$ & Input and denoised $i$-th latent sample at timestep $t$\\
        $t,T$ & Timestep, Max number of timesteps\\
        $n, d, s$ & Number of samples, real feature size, latent feature size\\
        $\beta^t, \bar{\alpha}^t$ & Variance schedule at timestep $t$, and $\prod_{j=1}^t(1-\beta^t)$\\
        $\theta, p_\theta, \epsilon_\theta$& Neural network parameters, reverse process, estimated noise\\
        $\mathcal{L}^t,\mathcal{M}^t$ & Continuous loss, multinomial loss at timestep $t$\\
        $\epsilon$ & Base noise level\\
        $C_i$ & $i$-th client\\
        $F(\_,t,\epsilon)$ & Forward process output after $t$ timesteps\\
        $X_i, Z_i$ & Original features, latent features of $i$-th client\\
        $\tilde{X_i}$ & Denoised original features of $i$-th client\\
        $\mathcal{G}, \mathcal{E}_i, \mathcal{D}_i$ & Denoising backbone, encoder, and decoder at $i$-th client\\
        $\tilde{Z},\tilde{Z}_i$ & Centralized denoised latents, Denoised latents for $C_i$\\

 \hline

    \end{tabular}
    
    \label{tab:notations}
\end{table}

%% file: icde_vfl_difuss/experiments.tex
\section{Experiments}
\label{sec:experiments}
We assess \algo's performance on multiple aspects. First, we evaluate its synthetic data quality against centralized and decentralized methods, considering its resemblance to original data and utility in downstream tasks. Second, we compare \algo's communication costs with end-to-end training, highlighting the reduced costs of stacked training. Third, we analyze \algo's privacy risks when sharing synthetic features post-generation. We also test \algo's robustness to feature permutation under a varying number of clients and partition sizes\footnote{Full experimental results available here: \footurl}.

\begin{figure}[tb]
    \centering
    \includegraphics[width=0.4\textwidth, angle=0]{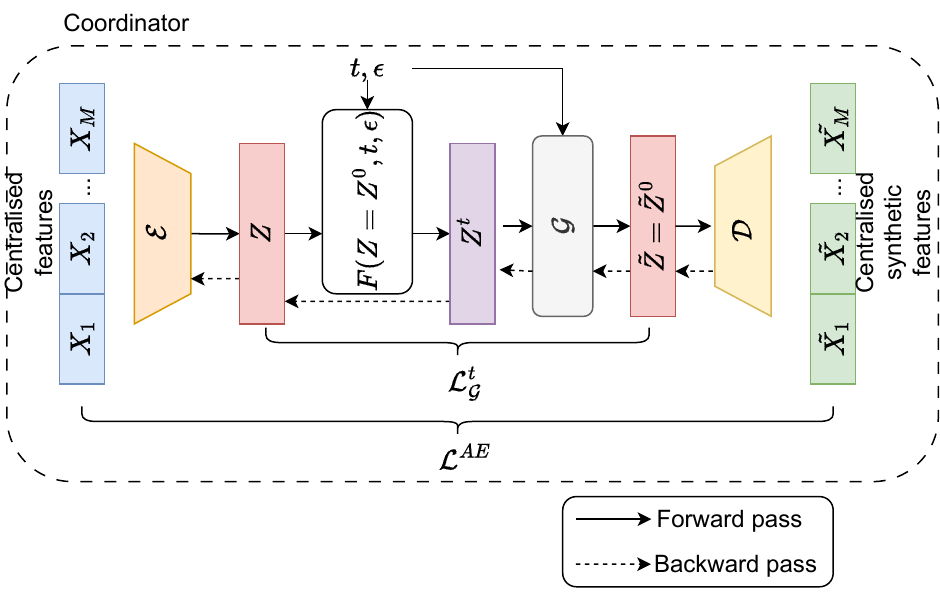}
    \caption{\algoe: centralized end-to-end baseline. }
    \label{fig:e2e}
\end{figure}
\begin{figure}[tb]
    \centering
\includegraphics[width=0.4\textwidth,angle=0]{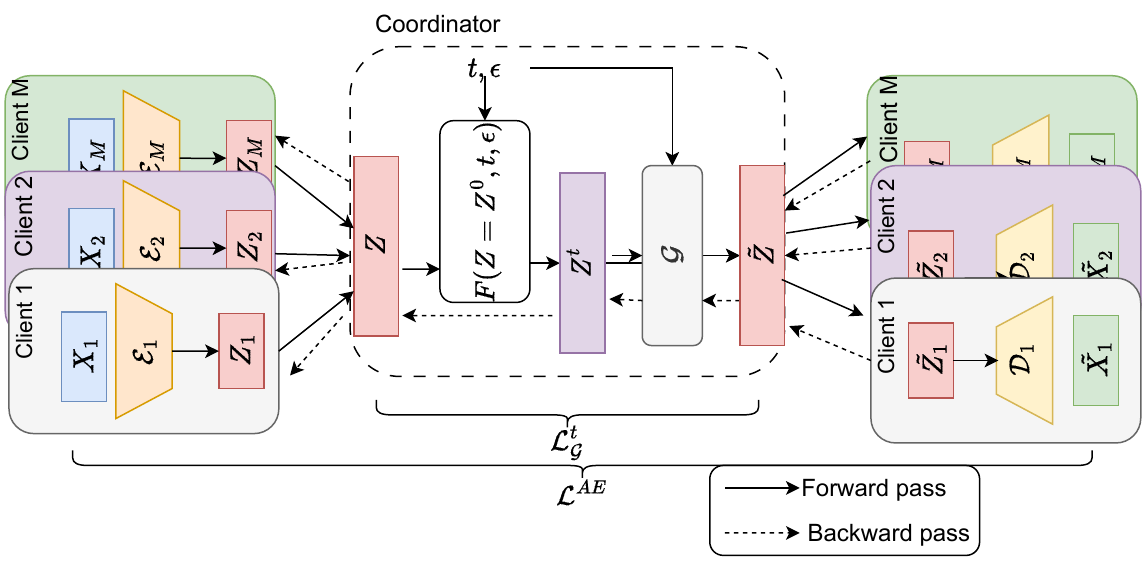}
    \caption{\algoed: distributed end-to-end baseline. Training Loss = $\mathcal{L}_\mathcal{G}^t + \mathcal{L}^{AE}$ }
    \label{fig:e2emulti}
\end{figure}

\subsection{Experiment Setup}
\label{s:setup}

\para{Hardware \& Software} We run experiments on an AMD Ryzen 9 5900 12-core processor and an NVIDIA 3090 RTX GPU with CUDA version 12.0. The OS used is Ubuntu 20.04.6 LTS. PyTorch is used to simulate the distributed experiments by creating multiple models corresponding to individual clients.

\para{Datasets}
To assess the quality of generated data, we use nine benchmark datasets used in generative modeling: Abalone~\cite{abalone}, Adult~\cite{adult}, Cardio~\cite{cardio}, Churn Modelling~\cite{churn}, Cover~\cite{cover}, Diabetes~\cite{Diabetes}, Heloc, Intrusion~\cite{intrusion}, and Loan~\cite{loan}. These are well-known benchmarks in the context of generative modeling~\cite{zhao2021ctab,kotelnikov2023tabddpm}.
 \input{icde_vfl_difuss/tables/datastats}

From \autoref{tab:DataStatistics}, we can classify the datasets into three categories based on the number of features: \textbf{i. Easy} - Abalone, Diabetes, Cardio. \textbf{ii. Medium} - Adult, Churn, Loan. \textbf{iii. Hard} - Intrusion, Heloc, Cover. \autoref{tab:DataStatistics} also includes the total feature sizes before and after one hot encoding and their corresponding changes (used by \algotab and GANs).

\para{Baselines} 
\revision{We tested centralized tabular GAN methods~\cite{zhao2021ctab,xu2019modeling,tablegan} falling under two architectural flavors: linear backbone, i.e., CTGAN (\galgol)~\cite{xu2019modeling}, and convolutional backbone, i.e., CTAB-GAN (\galgoc)~\cite{zhao2021ctab}, which extends CTGAN and Table-GAN~\cite{tablegan}}. For DDPMs, the models compared are tabular latent-diffusion models, \algo, and its centralized counterpart \algol. In addition, we also compare these models with their corresponding end-to-end trained versions, i.e., centralized \algoe (end-to-end centralized) and distributed \algoed (end-to-end distributed), with architectures as shown in Fig.~\ref{fig:e2e} and Fig.~\ref{fig:e2emulti} respectively. State-of-the-art \algotab is also compared.
\input{icde_vfl_difuss/tables/resutil}

The architectures of \algoe and \algoed, shown in Fig.~\ref{fig:e2e} and Fig.~\ref{fig:e2emulti} respectively, consist of an encoder followed by the DDPM and a decoder at the end. During training, the encoder(s) first compute(s) latents. The DDPM unit adds noise and iteratively denoises it to generate synthetic latents. The decoder(s) then recover the output in the original space and compute the losses individually using \eqref{eq:aeloss}. The MSE loss from the DDPM component is added \eqref{eq:smalldiff2}, and the entire network is trained end-to-end.

\para{Training configurations} For \algo, \algol, \algoe, and \algoed models, we utilize an autoencoder architecture with three linear layers for encoders and decoders. The activation function used is GELU~\cite{hendrycks2023gaussian}. In centralized versions, the embedding and hidden dimensions are set to 32 and 1024, respectively, equally partitioned between clients in the distributed versions. The latent dimension is set to the number of original features before one-hot encoding. In the case of distributed models (\algo, \algoed), the centralized autoencoders are evenly split across different clients. However, for \algotab, which lacks autoencoders, a 6-layer MLP with a hidden dimension of 256 forms the neural backbone of its DDPM. For the remaining models, a neural backbone for the DDPM consists of a bilinear model comprising eight layers with GELU activation and a dropout factor of 0.01. \revision{For the GAN baselines, we use four convolutional or linear layers with leaky ReLU activation and layer norm for the generator. The discriminator uses the transposed architecture but is otherwise similar.} All models are trained for 500,000 iterations encompassing the autoencoders and the DDPM's neural network, using a batch size 512 and a learning rate of 0.001. DDPM training involves a maximum of 200 timesteps, with inference conducted over 25 steps. In the case of distributed models (\algo and \algoed), dataset features are partitioned equally among four clients. The last client gets any remaining features post-division without shuffling.

\subsection{Benchmark Framework}

\label{subsec:benchmark}
\algo uses three types of metrics for the evaluation: resemblance, utility, and privacy.
All metrics are in the range of (0-100), with 100 being the ``best". The quality of synthetic data is calculated using resemblance and utility. Privacy metrics quantify the leakage associated with centralizing the synthetic features post-generation. We explain these metrics in detail as follows.

\parab{Resemblance}  measures how close the synthetic data is to the original data regarding feature distributions. It is a composite score that considers similarities between the real and synthetic data by computing the mean of the following five scores:
\begin{enumerate} 
\item Column Similarity: The correlation between the values in each real and synthetic column using the Pearson correlation~\cite{sedgwick2012pearson} for numerical columns, and Theil's U~\cite{bliemel1973theil} for categorical ones.
\item Correlation Similarity: The correlation between the correlation coefficients between each pair of columns using pairwise Pearson correlation and Theil's U for numerical and categorical features, respectively.
\item Jensen-Shannon Similarity~\cite{menendez1997jensen}: This computes the distance between the probability distributions of the real and synthetic columns. One minus the distance is used so that all metrics are comparable, with higher scores being better.
\item Kolmogorov-Smirnov Similarity~\cite{berger2014kolmogorov}: This distance measure computes the maximum difference between cumulative distributions of each real and synthetic feature. One minus the distance makes higher scores better.
\item Propensity Mean-Absolute Similarity~\cite{olmos2015propensity}: A binary classifier (XGBoost~\cite{chen2016xgboost}) is trained to discriminate between real and synthetic samples. When the classifier cannot distinguish between the two datasets, the mean-absolute error of its probabilities is 0. One minus the error is used so that higher scores are better.
\end{enumerate}

\parab{Utility} measures how the synthetic data performs on a downstream prediction task using an XGBoost model. Real or synthetic datasets are used for training, and evaluation is done using the same real hold-out set in each case. The downstream performance is calculated by taking the 90th percentile of macro-averaged F1 scores for categorical columns and the D2 absolute error scores for continuous columns. Finally, the utility score is calculated by taking the ratio of the synthetic to the real data's downstream performance (in percent, clipped at the max value of 100). 

\parab{Privacy}  
 quantifies the risk associated with sharing the synthetic features post-generation by averaging the scores from three attacks based on the framework in~\cite{giomi2022unified} and~\cite{houssiau2022tapas}. Again, higher scores indicate better resistance against the following three attacks.
\begin{enumerate}
\item \emph{Singling Out Attack}~\cite{giomi2022unified}: This singles out individual data records in the training dataset. If unique records can be identified, the synthetic data might reveal individuals based on their unique attributes.
\item \emph{Linkability Attack}~\cite{giomi2022unified}: This associates two or more records, either within the synthetic dataset or between the synthetic and original datasets, by identifying records linked to the same individual or group. 
\item \emph{Attribute Inference Attack}~\cite{houssiau2022tapas}: This deduces the values of undisclosed attributes of an individual based on the information available in the synthetic dataset.

\end{enumerate}





\subsection{Quantitative analysis}
\para{Setup}  We quantitatively analyze the resemblance and utility scores on the nine diverse datasets (abalone, adult, cardio, cover, churn\_modelling, diabetes, heloc, intrusion, and loan). \revision{The models compared are \algo, GANs (\galgol, \galgoc), centralized latent diffusion (\algol), end-to-end baselines (\algoe, \algoed), and \algotab.} Four clients were used for the distributed models (\algo and \algoed). The resemblance and utility scores (0-100) are averaged from 5 trials, results for which are shown in \autoref{tab:resemblance} and \autoref{tab:utility}.
\input{icde_vfl_difuss/tables/correlations}

\para{Comparable model performance to centralized methods} 
\revision{Coinciding with the results of other works~\cite{dhariwal2021diffusion,kotelnikov2023tabddpm}, we see that our decentralized latent diffusion method significantly outperforms centralized GANs. Improvements of up to \textbf{43.8} and \textbf{29.8} percentage points are achieved over GANs on resemblance and utility. This could be due to instabilities with training GANs leading to mode collapse~\cite{bayat2023study}. As a result, the generator may end up oversampling cases that escape the notice of the discriminator, leading to low diversity.} Within DDPMs, \algotab's and \algol's performance are generally an upper bound to what \algo can achieve since they are centralized baselines. Despite this, \algo achieves comparable resemblance and utility scores, especially compared to its centralized version \algol. Interestingly, the end-to-end versions \algoe and \algoed perform the worst. This may be because the DDPM backbone at the intermediate level needs to add noise to the latents and then denoise them during training. This may be problematic, especially during the initial training phases, as the received latents are already quite noisy since the autoencoders are initially untrained. Therefore, noise is added to more noise, making it difficult to distinguish the ground truth from actual noise. On the other hand, with \algo and \algol, the noise needs to be removed from fully-trained autoencoders. So, the latent values are less noisy, making it easier for the DDPM to distinguish between noise and actual latents during training correctly.


\para{Result analysis} We generally observe that latent DDPMs are better performing than \algotab on challenging datasets with a lot of sparse features, such as Intrusion and Cover, whereas \algotab does better on more straightforward datasets with a smaller number of features, such as Loan, Adult, and Diabetes. This indicates that combining multinomial and MSE loss may be better for high-quality data generation with low feature sizes and sparsity. At the same time, the conversion into latent space may be better for highly sparse datasets with large cardinality in discrete values.

\vspace{-1mm}
\subsection{Qualitative Analysis}

 \para{Setup} We qualitatively analyze the generated synthetic data based on the feature correlation differences between the real and synthetic features. Based on the resemblance and utility scores from \autoref{tab:resemblance} and \autoref{tab:utility}, we select the top three models (\algotab, \algol, and \algo) and show the correlation difference graphs for one of the simpler datasets (Cardio) and one hard dataset (Intrusion). Again, four clients were used for \algo\footnote{Additional results on feature distributions of real and synthetic data are available in the appendix: \footurlnew}. 
 
\para{High correlation resemblance of \algo} The graphs in \autoref{tab:correlation} show that \algo captures feature correlations well and is comparable to the centralized version, i.e., \algol's performance. It also performs better than \algotab on the harder dataset, Intrusion, as we see a darker shade for the correlation differences for \algotab. However, we observe that \algotab performs better on the simpler dataset, Cardio, due to the lower difference. These results match the scores indicated in \autoref{tab:resemblance} and \autoref{tab:utility}. As explained earlier, the large feature size of Intrusion makes it more challenging to model. It is amplified in difficulty for \algotab due to the additional sparsity induced by one-hot encoding. This makes it do worse than the latent space models. On Cardio, the problem of sparsity is not amplified by a lot as the number of discrete features is lower, enabling \algotab to perform better. Nevertheless, \algo does not perform much worse than \algotab even on Cardio, as the correlation differences are not very large.  



\input{icde_vfl_difuss/tables/privacy}

\subsection{Communication Efficiency}

\para{Setup} To demonstrate the benefits of stacked training over end-to-end training with increasing iterations, we compare \algo with \algoed, using the same parameter sizes for the autoencoders and the DDPM's neural backbone in each. Again, we use four clients with equally partitioned features. The iterations are varied between 50,000, 500,000, and 5 million. We measure the total bytes transferred between the clients and the coordinator over the course of the training process. Without loss of generality, we show the results on two datasets. One simple dataset with fewer features, i.e., Abalone, and one difficult dataset, Intrusion.

\para{Constant communication cost of \algo} From Fig.~\ref{fig:scalability}, we see that changing the number of iterations does not increase the communication cost of \algo. Due to the stacked training, the latents for the original training data only need to be transferred to the coordinator once after training autoencoders on each client. As the autoencoders and the DDPM are trained independently, there is no requirement to communicate gradients and forward activations between clients repeatedly, which results in a single round. \revision{End-to-end training methods such as \algotab, \algoe, and \algoed thus suffer from increasing costs as the iterations increase, i.e., they have cost $O(\#epochs)$. Naively distributing \algotab would incur even higher costs than \algoed, as one-hot encoding significantly increases the feature sizes being communicated (see \autoref{tab:DataStatistics}). As shown in the table, some datasets have a significant increase in size, leading to a proportionate rise that exceeds even \algoed: Churn (\textgreater200x), Heloc (\textgreater9x), Adult (\textgreater7x), and Intrusion (\textgreater6x).}

\input{icde_vfl_difuss/tables/scalability}

\begin{figure*}[tb]
\centering
\begin{subfigure}{0.3\textwidth}
    \includegraphics[width=\textwidth]{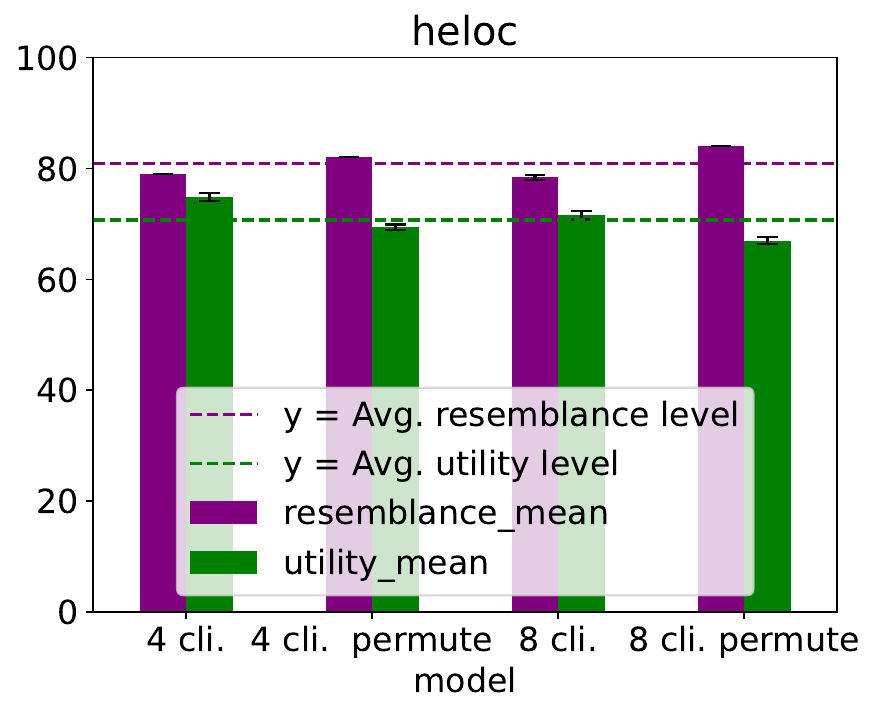}
    \caption{Heloc}
    \label{subfig:heloc_robust}
\end{subfigure}
\begin{subfigure}{0.3\textwidth}
    \includegraphics[width=\textwidth]{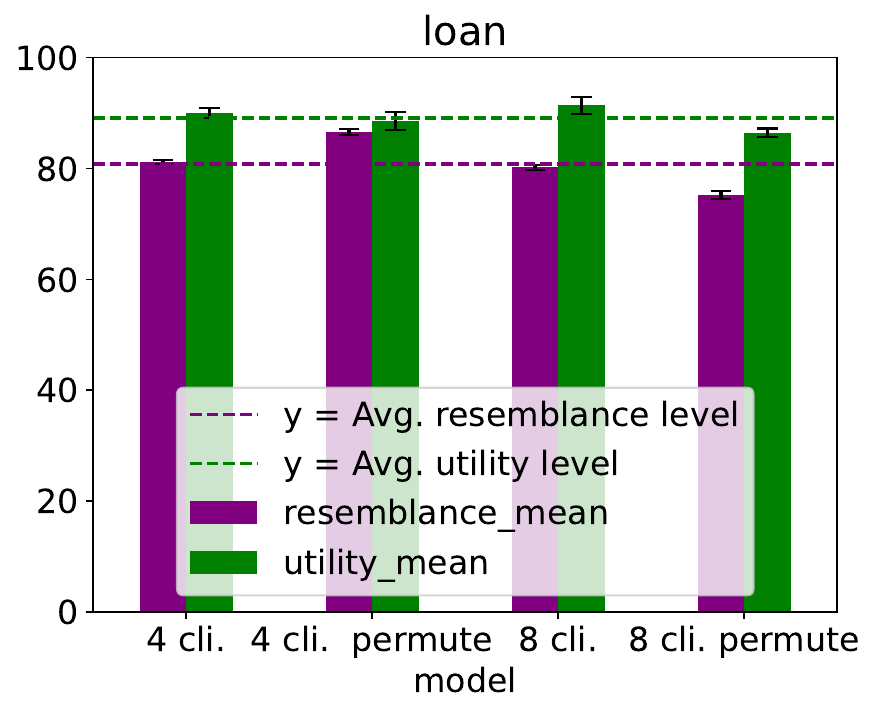}
    \caption{Loan}
    \label{subfig:loan_robust}
\end{subfigure}
\begin{subfigure}{0.3\textwidth}
    \includegraphics[width=\textwidth]{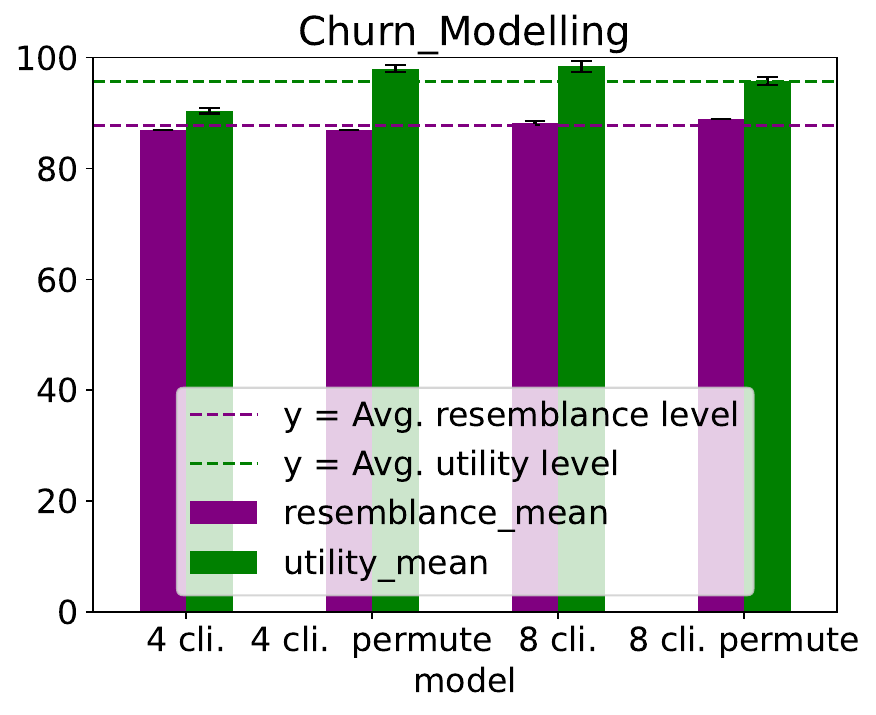}
    \caption{Churn}
    \label{subfigLchurn_robust}
\end{subfigure}
\caption{Robustness results for 4 and 8 clients with two different partitions (default and permuted)}
\label{fig:robustness}
\vspace{-3mm}
\end{figure*}

\subsection{Empirical Privacy Risk Analysis}
\label{sec:empiricalprivacy}
\para{Setup} Typically, post-generation, the synthetic data partitions from each client are shared with all parties. 
Although this methodology offers significant practical utility, it also poses privacy risks. Specifically, the distributions of the synthetic data could leak information on the original features, e.g., attribute inference attacks~\cite{houssiau2022tapas}. We quantify this risk by computing the privacy scores as explained in \hyperref[subsec:benchmark]{Section~V-F}. The best-performing methods, \algo. \algol, and \algotab are compared again, with the results shown in \autoref{tab:privacy}. 

\para{Improved privacy against centralized baselines} From the results, we see that \algo has the overall best privacy score, indicating it runs the lowest risk of leaking private feature information post-generation. While \algotab and \algol were able to achieve better resemblance and utility scores, we see that they lack in terms of privacy. \algo achieves higher privacy scores than the centralized version, \algol, on 8 out of 9 datasets, ranging from 4.5\% (Cardio, Heloc), rising to 18.4\%, 15.7\%, and 14.2\% on Adult, Diabetes, and Churn respectively.  

\para{Privacy-quality tradeoffs} Despite the higher privacy score of \algo, there is an inherent tradeoff between achieving very high utility and resemblance versus high privacy. For example, in Intrusion data, we observe that \algotab achieves the highest privacy score but has significantly lower resemblance and utility. Hence, data privacy may be better, but the synthetic samples may be unusable for the downstream task. Conversely, training the generative model using centralized features allows the model to capture relations between features better, leading to higher resemblance and utility. However, having synthetic features that highly resemble original features will enable adversaries to map the associations between features better, lowering privacy. While differential privacy adds noise to synthetic data for better privacy, it can lead to performance tradeoffs that are hard to control~\cite{wang2023differentially,tajeddine2020privacy}.

\revision{We also evaluate the sensitivity of privacy scores when the number of denoising steps varies. Two datasets were chosen: one easier dataset (abalone) and one difficult (heloc). As shown in \autoref{tab:sensitivity}, increasing the denoising steps allows the latent diffusion model to remove more noise, but lowers the privacy score. Notably, the privacy scores saturate quickly upon increasing the number of denoising steps. Hence, the scores lose sensitivity to the noise level within a few timesteps. }

\input{icde_vfl_difuss/tables/sensitivity.tex}



\subsection{Robustness to client data distribution}
\para{Setup} \algo's robustness to differing partition sizes on clients and permuted feature assignments is measured. We vary the clients between 4 and 8 and also experiment with two orders of the feature assignments to clients. The first order maintains unshuffled columns. The second order shuffles the columns using a seed value of 12343. After shuffling, we partition the columns for allocation to each client. Without loss of generality, we show the results (Fig.~\ref{fig:robustness}) for three datasets: Heloc, Loan, and Churn. Results on additional datasets are available in the link pointing to the full experimental results (see \hyperref[sec:experiments]{Section V}).

\para{Robustness to feature permutation and partitioning} 
From the figures, we see that changing the number of clients or shuffling the features does not make the resemblance/utility deviate too much from the average level, indicating that \algo is generally invariant to feature permutation. This is because centralization of the latents at the DDPM allows it to capture cross-feature correlations that may have been missed or poorly captured at the local encoders. This enables the model to achieve similar resemblance and utility scores.

Exceptions stand out in specific scenarios. For instance, resemblance notably drops in the Loan dataset transitioning from 4 to 8 clients with permuted feature assignments (Fig.~\ref{subfig:loan_robust}). In the Churn dataset, utility significantly improves when moving from default to permuted partitioning (Fig.~\ref{subfigLchurn_robust}). The drop in resemblance might arise from correlated features being reassigned to different clients in the 8-client setup, delaying the learning of associations. Improved performance for the latter case may stem from separating jointly-existing noisy features during re-partitioning or permutation.


 


%% file: icde_vfl_difuss/tables/datastats.tex
\begin{table}[t]
\centering
\caption{Statistics of Datasets. We provide the number of rows, categorical and numerical features, the total size before (\#Bef) and after (\#Aft) one hot encoding, and the increase in feature size (Incr).}
\begin{tabular}{cccc*{3}{>{\color{black}}c}}
\Xhline{1.5pt}
        & \#Rows & \#Cat. & \#Num. &\#Bef. & \#Aft. & Incr. \\ \hline
Loan    & 5000     & 7& 6 & 13 & 23 & \textbf{1.77x}          \\ \hline
Adult   & 48842    & 9& 5 & 14 & 108 & \textbf{7.71x}          \\ \hline
Cardio  & 70000    & 7  & 5 & 12 & 21 & \textbf{1.75x}              \\ \hline
Abalone & 4177 & 2 & 8 & 10 & 39 & \textbf{3.9x} \\\hline
Churn & 10000&8 &6 &14 & 2964 & \textbf{211.71x}\\\hline
Diabetes & 768 &2 & 7 &9 & 26 & \textbf{2.89x} \\\hline
Cover & 581012&45 &10 &55 &104 & \textbf{1.89x}\\\hline
Intrusion &22544 &22 &20 &42 &268 & \textbf{6.38x}\\\hline
Heloc &10250 &12 &12 &24 &239 & \textbf{9.96x}\\

\Xhline{1.5pt}

\end{tabular}

\label{tab:DataStatistics}
   \vspace{-3mm}
\end{table}

%% file: icde_vfl_difuss/tables/resutil.tex
\begin{table*}[bt]
    \centering
    \caption{Resemblance Scores (0-100). Higher scores indicate better resemblance. Zero std. deviation indicates negligible deviation. Percentage point difference (PPD) of \algo with the best GAN is also shown.}
    \scalebox{0.9}{
    \begin{tabular}{c c c c c c c c c c}
    \Xhline{1.5pt}
       Model& Abalone & Adult & Cardio & Churn & Cover & Diabetes & Heloc & Intrusion & Loan\\\hline

       \rowcolor{white} 
       \galgoc & 64.0 $\pm$ 0.00 & 38.0 $\pm$ 0.00 & 59.8 $\pm$ 0.40 & 43.4 $\pm$ 0.80 & 45.2 $\pm$ 0.40 & 75.8 $\pm$ 0.40 & 54.0 $\pm$ 0.00 & 47.2 $\pm$ 0.40 & 76.4 $\pm$ 0.49\\ 

       \rowcolor{white}
       \galgol & 54.2 $\pm$ 0.40 & 28.6 $\pm$ 0.49 & 29.0 $\pm$ 0.00 & 30.8 $\pm$ 0.39 & 36.0 $\pm$ 0.00 & 51.0 $\pm$ 0.00 & 48.0 $\pm$ 0.00 & 39.0 $\pm$ 0.00 & 40.0 $\pm$ 0.00 \\

       \algoe & 85.2$\pm$0.40 &  60.0$\pm$0.00 &  60.2$\pm$0.40 &  88.2$\pm$0.40 &  51.0$\pm$0.40 &  72.4$\pm$0.80 &  68.4$\pm$0.49 &  48.0 $\pm$0.00 &  81.2$\pm$0.40\\

       \algoed & 56.4$\pm$0.80 &  46.0$\pm$1.09 &  44.0$\pm$0.89 &  78.0$\pm$0.00 &  40.8$\pm$0.40 &  61.8$\pm$ 3.12 &  61.0$\pm$0.00 &  37.0$\pm$0.00 &  49.8$\pm$1.16\\

        \algotab  & 91.2$\pm$0.75 & \textbf{97.0$\pm$0.00} & \textbf{98.0$\pm$0.00} & 63.6$\pm$0.49 & 78.0$\pm$0.00 & \textbf{94.6$\pm$0.49}& \textbf{88.0$\pm$0.00} &44.0$\pm$0.00 &\textbf{98.0$\pm$0.00}\\

        \algol & \textbf{92.0$\pm$0.00} &  78.0$\pm$0.00 &  72.2$\pm$0.40 &  \textbf{89.0$\pm$0.00} &  \textbf{92.0$\pm$0.00}&  90.0$\pm$0.63 &  83.4$\pm$0.49 &  \textbf{68.0$\pm$0.00}&  83.4$\pm$0.49\\
        
        \algo & 91.0$\pm$0.00 &  73.0$\pm$0.00 &  71.0$\pm$0.00 &  87.0$\pm$0.00 &  89.0 $\pm$ 0.00 &  84.0$\pm$0.63 &  79.0$\pm$0.00 &  67.0$\pm$0.00 &  81.2$\pm$0.40\\\hline
        \rowcolor{white}
        PPD (vs GAN)&27.0 &35.0&11.2&43.6&43.8&8.2&25.0&19.8&4.8\\
        \Xhline{1.5pt}
    \end{tabular}}
    
    \label{tab:resemblance}
\end{table*}

\begin{table*}[tb]
    \centering
    \caption{Utility Scores (0-100). Higher scores indicate better downstream utility. Percentage point difference (PPD) of \algo with the best GAN is also shown.}
    \scalebox{0.9}{
    \begin{tabular}{c c c c c c c c c c}
    \Xhline{1.5pt}
       Model& Abalone & Adult & Cardio & Churn & Cover & Diabetes & Heloc & Intrusion & Loan\\\hline

       \rowcolor{white}
       \galgoc & 71.0 $\pm$ 0.63 & 82.6 $\pm$ 11.30 & 94.0 $\pm$ 3.63 & 85.0 $\pm$ 1.09 & 82.6 $\pm$ 1.01 & 96.2 $\pm$ 2.56 & 45.0 $\pm$ 0.63 & 36.2 $\pm$ 0.98 & 82.6 $\pm$ 0.49\\

       \rowcolor{white}
       \galgol & 65.2 $\pm$ 0.40 & 30.6 $\pm$ 1.62 & 47.6 $\pm$ 0.49 & 78.4 $\pm$ 1.02 & 36.2 $\pm$ 0.40 & 84.6 $\pm$ 1.62 & 38.4 $\pm$ 0.49 & 25.4 $\pm$ 0.49 & 82.0 $\pm$ 0.63\\

       \algoe & 70.0$\pm$1.41 &  45.0$\pm$1.67 &  75.2$\pm$1.47 &  87.8$\pm$0.74 &  89.8$\pm$0.40 &  84.2$\pm$5.11 &  39.8$\pm$0.40 &  31.0$\pm$0.63 &  77.0$\pm$0.89\\
       
       \algoed & 70.8$\pm$1.72 &  33.6$\pm$1.02 &  55.3$\pm$0.33 &  87.0$\pm$0.89 &  56.8$\pm$1.46 &  89.0$\pm$2.28 &  39.2$\pm$0.40 &  24.2$\pm$1.16 &  71.2$\pm$2.92\\

       \algotab& 98.4$\pm$0.49 &89.2$\pm$1.60 & \textbf{99.6$\pm$0.49} &44.4$\pm$6.28 &80.4$\pm$7.94 &\textbf{100.0$\pm$0.00}&\textbf{96.4$\pm$0.49} &23.0$\pm$4.98 &\textbf{98.8$\pm$1.17}\\
       
        \algol & \textbf{100.0$\pm$0.00}&  \textbf{100.0$\pm$0.00} &  86.4$\pm$2.33 &  \textbf{100.0$\pm$0.00}&  95.8$\pm$0.40 &  99.6$\pm$0.49 &  76.4$\pm$0.49 &  61.2$\pm$1.16 &  94.8$\pm$1.93\\

        \algo & 97.2$\pm$1.16 &  96.6$\pm$5.04 &  93.2$\pm$4.95 &  90.4$\pm$0.49 & \textbf{96.4$\pm$0.49}&  95.2$\pm$3.92 &  74.8$\pm$0.74 &  \textbf{64.2$\pm$0.74}&  90.0$\pm$0.89\\\hline

        \rowcolor{white}
        PPD (vs GAN)&26.2&14.0&-0.8&5.4&13.8&-1.0&29.8&28.0&7.4\\
        \Xhline{1.5pt}
    \end{tabular}}
    
    \label{tab:utility}
\end{table*}

%% file: icde_vfl_difuss/tables/correlations.tex
\begin{table}[bt]
\centering
\caption{Feature correlation differences between real and synthetic data. Darker colour indicates worse performance.}
    \begin{tabular}{c c c c }
       & \algo  &  \algol & \algotab\\
        \begin{sideways}
            Cardio
        \end{sideways}&\includegraphics[width=0.13\textwidth]{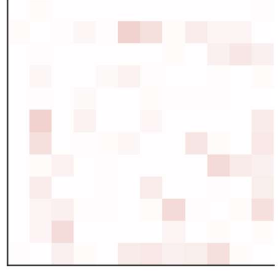}& \includegraphics[width=0.13\textwidth]{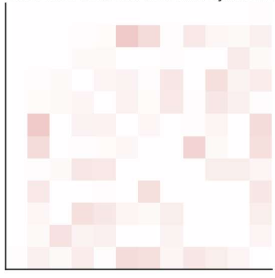}& \includegraphics[width=0.13\textwidth]{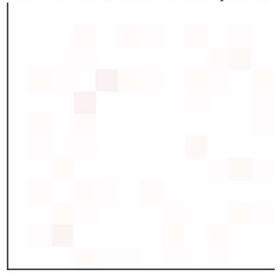}\\
        
        \begin{sideways} Intrusion \end{sideways} & \includegraphics[width=0.13\textwidth]{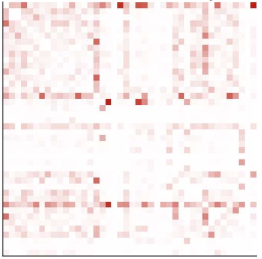}&\includegraphics[width=0.13\textwidth]{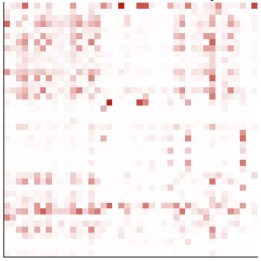}& \includegraphics[width=0.13\textwidth]{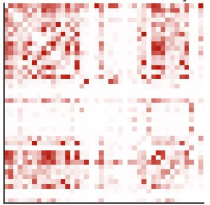} \\
        
    \end{tabular}
    
    \label{tab:correlation}
\end{table}

%% file: icde_vfl_difuss/tables/privacy.tex
\begin{table*}[tb]
    \centering
    \caption{Privacy scores of latent models and TabDDPM}
    \scalebox{0.9}{
    \begin{tabular}{c c c c c c c c c c}
        \Xhline{1.5pt}
          Model & Abalone & Adult & Cardio & Churn & Cover & Diabetes & Heloc & Intrusion & Loan  \\\hline

         \algotab & 48.2$\pm$0.48 & 70.1$\pm$2.61 &76.1$\pm$2.19 &86.7$\pm$0.24 &59.1$\pm$3.85 &46.2$\pm$0.69 & \textbf{56.7$\pm$1.60}&\textbf{92.1$\pm$2.62}&57.9$\pm$1.87\\
        
         \algol & 50.3 $\pm$ 0.58 &73.7 $\pm$ 2.29 &88.7 $\pm$ 2.82 &78.1 $\pm$ 4.18 &55.7 $\pm$ 1.81 &62.4 $\pm$ 1.32 &51.9 $\pm$ 1.46 &70.5 $\pm$ 2.89 &64.8 $\pm$ 2.36\\

         \algo &  \textbf{55.9$\pm$0.46} &\textbf{92.1$\pm$0.60} &\textbf{93.2$\pm$4.97}&\textbf{92.3$\pm$1.84}&\textbf{65.1$\pm$2.25}&\textbf{78.1$\pm$2.40}&56.4$\pm$1.58 &70.5$\pm$2.46 &\textbf{79.3$\pm$5.35}\\
        \Xhline{1.5pt}
    \end{tabular}}
    
    \label{tab:privacy}
\end{table*}

%% file: icde_vfl_difuss/tables/scalability.tex
\begin{figure}[tb]
    \centering
    \begin{subfigure}{0.23\textwidth}
    \includegraphics[width=\textwidth]{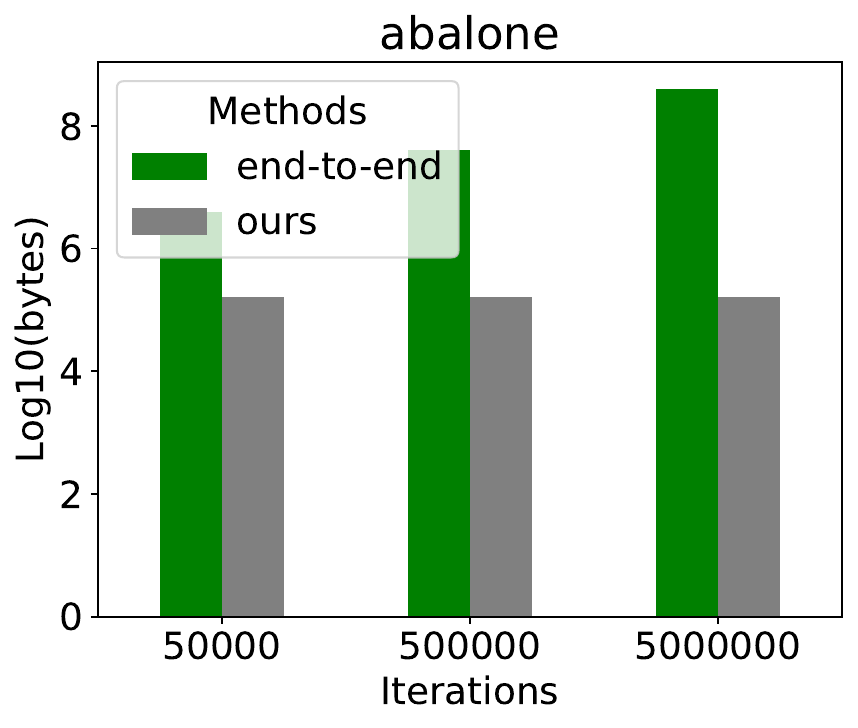}
    \end{subfigure}
    \begin{subfigure}{0.23\textwidth}\includegraphics[width=\textwidth]{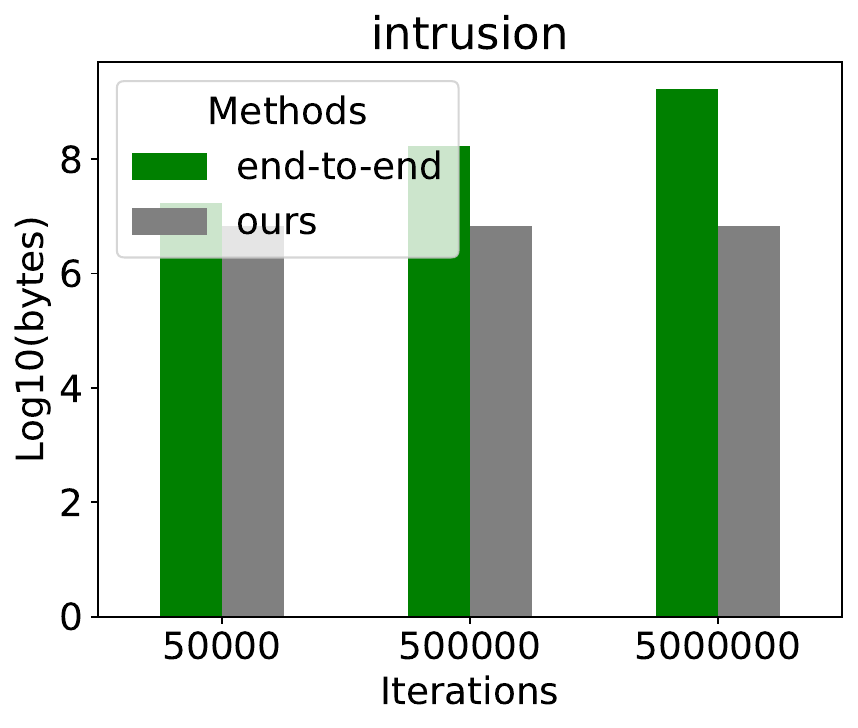}    
    \end{subfigure}    
    
    \caption{Bytes sizes communicated during training. \algo vs. \algoed}
    \label{fig:scalability}
    \vspace{-3mm}
\end{figure}

%% file: icde_vfl_difuss/tables/sensitivity.tex
\begin{table}[t]
    \centering
    \caption{Sensitivity of privacy score to the number of denoising steps}
    \color{black}
    \begin{tabular}{c c c c}
    \Xhline{1.5pt}
      \multirow{2}{*}{Dataset}   &  \multicolumn{3}{c}{Inference timesteps} \\
      & 2& 5 & 25\\
      \cline{1-4}\\
      Abalone & 58.15 $\pm$ 0.90 & 51.6 $\pm$ 0.41 & 50.3 $\pm$ 0.58\\
      Heloc & 66.1 $\pm$ 1.15 & 53.4 $\pm$ 0.56 & 51.9 $\pm$ 1.46 \\
    \Xhline{1.5pt}
    \end{tabular}
    
    \label{tab:sensitivity}
\end{table}

%% file: icde_vfl_difuss/relatedwork.tex
\begin{table}[tb]
    \caption{Overview of the related work. A yellow tick indicates partial satisfaction.}
    \footnotesize
    \scalebox{0.83}{
    \begin{tabular}{ccccc}
    \Xhline{1.5pt}
        Method & Type & Cross-silo & Tabular& Generating space\\\hline

        GTV~\cite{zhao2023gtv}&GAN &\tealcheck &\tealcheck &Real\\

        DPGDAN~\cite{wang2023differentially}&GAN &\tealcheck &\tealcheck &Real\\

        MedGAN~\cite{choi2017generating}& GAN &\redcross &\tealcheck &Latent\\

        VQ-VAE~\cite{oord2017neural} & VAE &\redcross &\tealcheck &Real\\

        TVAE~\cite{xu2019modeling} &VAE &\redcross &\tealcheck &Real\\

        dpart~\cite{mahiou2022dpart}& AR &\redcross &\tealcheck &Real\\

        DP-HFLOW~\cite{lee2022differentially} & Flow & \redcross &\tealcheck &Real\\

        STaSY~\cite{kim2022stasy}& Score-matching & \redcross &\tealcheck &Real\\

        Hoogeboom et al.~\cite{hoogeboom2021argmax} &DDPM/Flow & \redcross &\orangecheck& Real\\
        \algotab~\cite{kotelnikov2023tabddpm}& DDPM & \redcross & \tealcheck &Real\\

        Rombach et al.~\cite{rombach2022high}& DDPM &\redcross &\redcross &Latent\\

        \algo & DDPM & \tealcheck &\tealcheck &Latent\\

        \Xhline{1.5pt}
        
    \end{tabular}}
    
    \label{tab:relwork}
       \vspace{-3mm}
\end{table}

\section{Related Work}
\para{Tabular Synthesizers}
In the domain of generative modeling, deep neural network methods span variational autoencoders (VAEs), GANs, energy-based models (DDPMs, Score-matching), autoregressive models (AR), and flow-based methods, outlined in Bond et al.~\cite{bond2021deep}. Given our focus on tabular data synthesis, our exploration centers on the techniques delineated in \autoref{tab:relwork}, predominantly comprising tabular synthesizers. Although Hoogeboom et al.'s approach~\cite{hoogeboom2021argmax} pioneers multinomial DDPMs, it exclusively applies to categorical features, addressing only part of the tabular data synthesis need. Additionally, Rombach et al.'s work~\cite{rombach2022high}, a DDPM-based method inspiring \algo's latent space modeling, is tailored exclusively for image data.

\para{Cross-silo synthesis} \revision{Other than \algo, GAN-based methods, such as GTV~\cite{zhao2023gtv} and DPGDAN~\cite{wang2023differentially}, tackle the cross-silo challenge. Their backbones use centralized GAN architectures, i.e., Table-GAN, CTGAN, and CTAB-GAN~\cite{tablegan,xu2019modeling,zhao2021ctab}, which \algo outperforms. They also employ end-to-end training, increasing communication costs.}

\para{End-to-end (real space) vs. stacked training (latent space)} Most generative methods directly operate in the original/real space, requiring end-to-end training. In contrast, latent-based models can decouple training into local autoencoder training followed by generative modeling in the latent space. Limited work exists in this paradigm: MedGAN~\cite{choi2017generating} uses an autoencoder to transform features into continuous latents, followed by a GAN-based generative step, but it is centralized and unfit for cross-silo scenarios. Rombach et al.~\cite{rombach2022high} operates in the latent space solely for image data and in a centralized setup.

%% file: icde_vfl_difuss/futureworkandconclusion.tex
\section{Conclusion}
 We introduce \algo, a distributed latent DDPM framework for cross-silo tabular data synthesis. \revision{\algo trains models collaboratively by encoding original features into a latent space, ensuring confidentiality while avoiding the high costs of mainstream one-hot encoding.} We introduce a stacked training paradigm to decouple the training of autoencoders and the diffusion generator. This minimizes communication costs by transmitting latent features only once and avoids the risks of gradient leakage attacks. \revision{Centralizing latent features preserves cross-silo links while thwarting data reconstruction under vertically partitioned synthesis.}

\revision{Quantitative and qualitative analyses show \algo's comparable performance with centralized baselines regarding synthetic data quality, with diffusion-based models outperforming GANs. \algo achieves improvements of up to 43.8 and 29.8 percent points higher than centralized GANs on resemblance and utility, respectively. Stacked training reduces communication rounds to one step, significantly better than end-to-end training paradigms ($O(\#epochs)$).} When sharing synthetic data post-generation, \algo exhibits superior resistance against attacks, with minor trade-offs in data quality. Although robust under varied scenarios, slight susceptibility to feature permutation suggests room for future improvement.

Sharing synthetic features across parties poses challenges and potential privacy vulnerabilities. Future research could explore controlled information exchange or employ trusted third-party interventions to ensure privacy in collaborative training. Exploring methods like vertical federated learning would offer ways to achieve both privacy and high performance in downstream tasks by retaining feature-partitioning.

\section*{Acknowledgment}
This paper is supported by the project \textit{Understanding Implicit Dataset Relationships for Machine Learning} (VI.Veni.222.439), of the research programme NWO Talent Programme Veni, partly financed by the Dutch Research Council (NWO). It is also supported by the \textit{DepMAT} project (P20-22) of the NWO Perspectief programme.



